\documentclass[acmsmall,screen]{acmart}
\usepackage{times}
\usepackage{epsfig}
\usepackage{graphicx}
\usepackage{amsmath}
\usepackage{url}
\usepackage{amsfonts}
\usepackage{xcolor}
\usepackage{bm}
\usepackage{multirow}
\usepackage{colortbl}
\usepackage{hhline}
\usepackage{pifont}
\usepackage{mathtools}
\usepackage{wrapfig}
\usepackage[caption=false]{subfig}
\usepackage{ragged2e}
\usepackage{makecell}
\usepackage{enumitem}
\usepackage{float}
\usepackage{booktabs}
\usepackage{arydshln}

\AtBeginDocument{%
  }

\setcopyright{acmlicensed}
\copyrightyear{2018}
\acmYear{2018}
\acmDOI{XXXXXXX.XXXXXXX}

\acmJournal{CSUR}
\acmVolume{37}
\acmNumber{4}
\acmArticle{111}
\acmMonth{8}

\graphicspath{{./Imgs/},{./Imgs/Task_Pre/},{./Imgs/Authors/}}
\DeclareGraphicsExtensions{.pdf,.jpg,.png}

\newcommand{\para}[1]{\vspace{.04in}\noindent\textbf{#1}\quad}
\newcommand{\paraem}[1]{\vspace{.04in}\noindent\textit{#1}\quad}

\makeatletter
\g@addto@macro\normalsize{%
  \setlength\abovedisplayskip{3pt plus 2pt minus 1pt}%
  \setlength\belowdisplayskip{3pt plus 2pt minus 1pt}%
  \setlength\abovedisplayshortskip{2pt plus 1pt}%
  \setlength\belowdisplayshortskip{2pt plus 1pt}}
\makeatother

\newcommand{\gray}[1]{\textcolor[rgb]{0.7,0.7,0.7}{#1}}
\newcommand{\revise}[1]{#1}

\newcommand{\figref}[1]{Fig.~\ref{#1}}
\newcommand{\tabref}[1]{Tab.~\ref{#1}}
\newcommand{\secref}[1]{\S\ref{#1}}

\def\eg{\emph{e.g.}}
\def\vs{\emph{vs.~}}

\def\etal{\emph{et al.~}}

\begin{document}

\title{A Comprehensive Survey on Video Scene Parsing: Advances, Challenges, and Prospects}


\author{Guohuan Xie}
\affiliation{%
  \institution{College of Software, Nankai University}
  \city{Tianjin}
  \country{China}}
\email{xieguohuan@mail.nankai.edu.cn}
\orcid{0009-0009-6382-5302}

\author{Syed Ariff Syed Hesham}
\affiliation{%
  \institution{School of Electrical and Electronic Engineering, NTU \& Institute for Infocomm Research, A*STAR}
  \country{Singapore}}
\email{syedhesh002@e.ntu.edu.sg}
\orcid{0009-0000-9589-3056}

\author{Bing Li}
\affiliation{%
  \institution{School of Information and Communication Engineering, UESTC}
  \city{Chengdu}
  \country{China}}
\email{windof47@gmail.com}
\orcid{0000-0002-1875-2919}

\author{Wenya Guo}
\email{guowenya@dbis.nankai.edu.cn}
\orcid{0000-0001-5609-194X}
\author{Guolei Sun}
\email{guolei.sun@nankai.edu.cn}
\orcid{0000-0001-8667-9656}
\affiliation{%
  \institution{College of Computer Science, Nankai University}
  \city{Tianjin}
  \country{China}
}

\author{Ming-Ming Cheng}
\email{cmm@nankai.edu.cn}
\orcid{0000-0001-5550-8758}
\author{Yun Liu}
\authornote{Corresponding author: Yun Liu (E-mail: liuyun@nankai.edu.cn)}
\email{liuyun@nankai.edu.cn}
\orcid{0000-0001-6143-0264}
\affiliation{%
  \institution{VCIP, College of Computer Science, Nankai University}
  \city{Tianjin}
  \country{China}
}
\affiliation{%
  \institution{Academy for Advanced Interdisciplinary Studies, Nankai University}
  \city{Tianjin}
  \country{China}
}
\affiliation{%
  \institution{Nankai International Advanced Research Institute}
  \city{Shenzhen Futian}
  \country{China}
}

\renewcommand{\shortauthors}{Xie et al.}

\begin{abstract}
{Video Scene Parsing (VSP) studies dense video understanding, where every pixel in each frame must be segmented, each region must be named, and each object identity must remain coherent over time. This survey reviews recent progress in VSP across five tasks, spanning Video Semantic Segmentation (VSS), Video Instance Segmentation (VIS), Video Panoptic Segmentation (VPS), Video Tracking \& Segmentation (VTS), and Open-Vocabulary Video Segmentation (OVVS). We organize the literature as one architectural arc, running from hand-crafted motion and appearance cues through fully convolutional, attention-based and query-based designs to recent foundation-model approaches, and we trace how each family models temporal context, preserves identity, and balances accuracy against efficiency. We then compare the datasets, metrics and benchmark trends that shape current evaluation. Beyond cataloguing methods, we foreground the design trade-offs and recurring failure modes that cut across the field, namely temporal flicker, occlusion-induced identity switches, long-tail categories, and the annotation--capacity--latency tension. We close with open directions towards robust, efficient and open-world VSP systems.}
\end{abstract}




\maketitle

\begin{figure}[!h]
  \centering
  \includegraphics[width=0.9\linewidth]{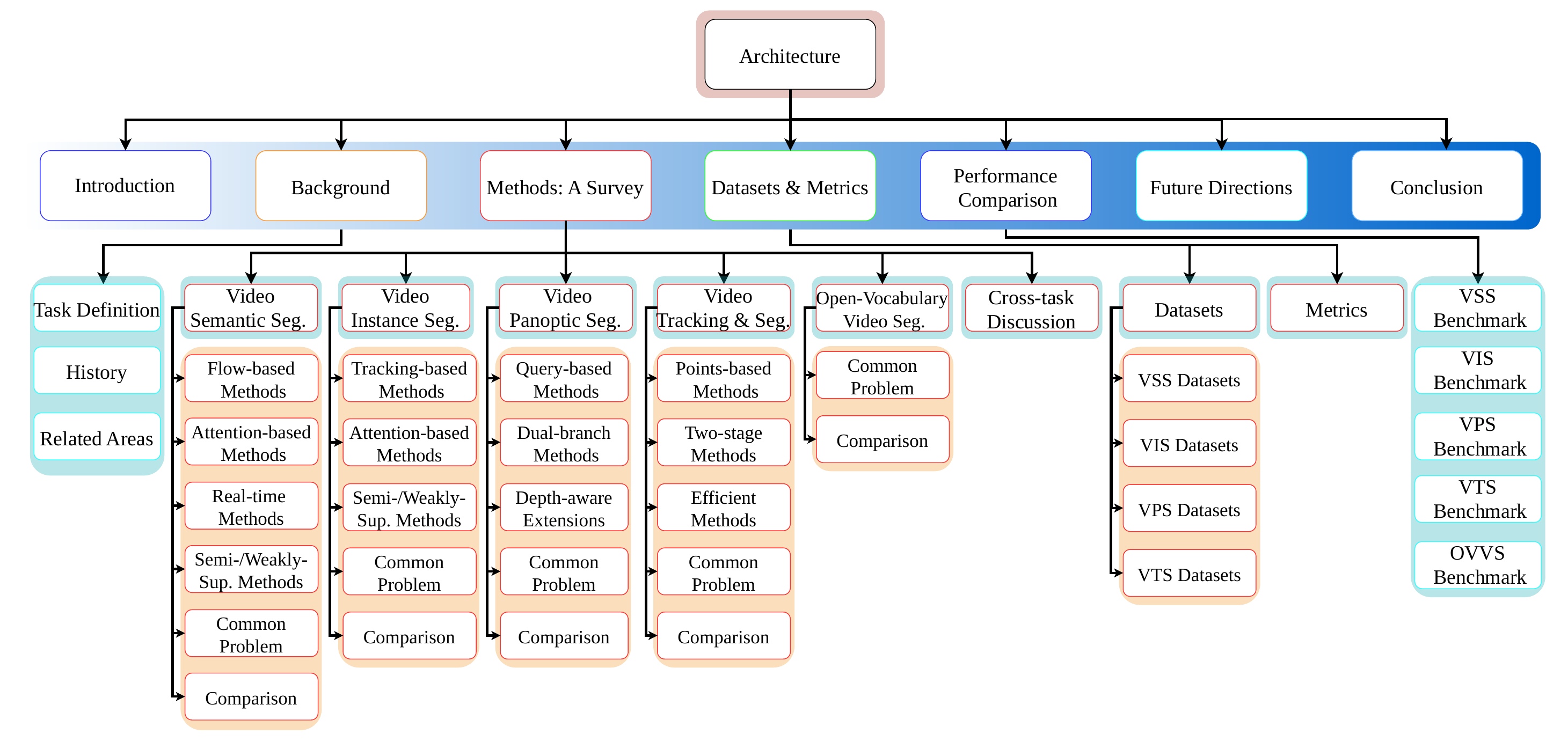}
  \vspace{-25pt}
  \caption{{\textbf{Overview of this survey.}}}
  \label{tab:structure}
  \vspace{-15pt}
\end{figure}

\section{Introduction}\label{sec:introduction}
{Video Scene Parsing (VSP) is a fundamental computer vision problem that assigns semantic labels, instance masks, and persistent identities to video pixels, depending on the task. It covers Video Semantic Segmentation (VSS), Video Instance Segmentation (VIS), Video Panoptic Segmentation (VPS), Video Tracking \& Segmentation (VTS), and Open-Vocabulary Video Segmentation (OVVS). By linking static image segmentation~\cite{shelhamer2017fully} with dynamic scene understanding~\cite{zhu2017deep}, VSP matters as a research problem and a practical perception module.
On the research side, it raises questions about temporal consistency~\cite{Gadde2017SemanticVC,Jain2018AccelAC,kundu2016feature}, spatio-temporal feature extraction~\cite{lee2021gsvnet,Ren2015FasterRT}, and robust tracking of dynamic objects~\cite{caelles20182018}. On the applied side, across autonomous driving, intelligent surveillance, robotics, and video editing, reliable VSP is the precondition for safe downstream decisions.}

{The machinery behind VSP has turned over three times. Early systems leaned on hand-crafted descriptors, among them color histograms, texture, and optical flow~\cite{lam1998video,fazekas2009dynamic,rashwan2013illumination}, paired with classical learners such as clustering~\cite{Yu2015EfficientVS}, graph-based methods~\cite{grundmann2010efficient}, SVMs~\cite{Perazzi2015FullyCO}, random forests~\cite{Badrinarayanan2013SemiSupervisedVS}, and Markov or conditional random fields~\cite{Jang2016StreamingVS,Liu2015MulticlassSV}. These pipelines were important foundations, yet they scaled poorly and demanded heavy domain-specific feature engineering. Deep learning broke that ceiling, since Fully Convolutional Networks (FCNs)~\cite{shelhamer2017fully,chen2018deeplab,zhao2017pyramid,ding2018context} learned hierarchical features for dense pixel-level prediction, and FCN-based methods~\cite{tsai2016video,zhu2019improving,huang2018efficient,paul2020efficient,zhao2018icnet} soon became the default framework across VSP. Transformers~\cite{vaswani2017attention} then opened the third era, because self-attention captures long-range dependencies and global context, and its vision variants~\cite{zhai2022scaling,meng2022adavit,liu2024vision,liu2022swin,chen2022scaling,yan2022multiview,gu2022multi,ding2024unireplknet,wu2022p2t,dosovitskiy2021image,Carion2020EndtoEndOD} have reshaped image and video segmentation by modeling spatial and temporal interactions beyond the locality of CNNs.}

{These methodological shifts have also widened the kinds of problems VSP is expected to solve. Video Tracking \& Segmentation (VTS) asks for more than per-frame segmentation, adding consistent identity preservation across frames~\cite{voigtlaender2019mots,cheng2023tracking}, which demands robust association under occlusion, abrupt motion and complex interactions, essential for multi-object tracking in crowded scenes and advanced editing pipelines. A further frontier, Open-Vocabulary Video Segmentation (OVVS), pairs CLIP-style vision--language priors~\cite{radford2021learning} with segmentation to escape the closed label sets of classical VSS. Recent open-vocabulary approaches~\cite{han2023global,zhang2023simple,xu2023learning,han2023open,chen2023open} segment objects beyond predefined categories, and generalize zero-shot to the rare or novel entities that pervade unconstrained video.}

\revise{Against this backdrop, existing surveys still leave three structural gaps. \textbf{First, fragmented scope.} The early survey of Wang \etal~\cite{Wang2021ASO} is anchored on Video Object Segmentation (VOS), treating VSS, VIS, and VPS only as adjacent topics, since VTS and OVVS were still nascent at its time of writing. Li \etal~\cite{Li2023TransformerBasedVS} focus narrowly on the Transformer paradigm, leaving the convolutional methodology that long sustained the field under-discussed. Other surveys either generalize from image segmentation~\cite{garcia2018survey,jiao2021new} or address open-vocabulary recognition in isolation~\cite{wu2024towards,zhu2024survey}, so no single survey covers the full VSP family. \textbf{Second, limited analysis.} Existing surveys largely follow a method-catalogue convention, listing models and accuracy numbers but rarely examining how competing methodological lineages shift across task settings, data scales, and compute budgets. They also under-analyze how four temporal-modeling mechanisms, namely flow warping, attention, query-based set prediction, and spatio-temporal memory, trade off accuracy, latency, memory footprint, and long-horizon modeling. \textbf{Third, missing cross-task synthesis.} VSS, VIS, VPS, VTS, and OVVS share the same core challenges of temporal consistency, identity preservation, long-tail recognition, and label efficiency, yet most surveys discuss them separately rather than through shared cross-task dimensions. This survey is built to connect these issues across tasks.}

\revise{Concretely, this survey makes three contributions. \textbf{First, a unified scope.} We treat VSP as one coherent family of \emph{five} tasks, namely VSS, VIS, VPS, VTS, and OVVS. \secref{sec:method} sets out each task's native methodological taxonomy, such as Flow-based, Attention-based, Real-time, and Semi-/Weakly-supervised VSS, while \secref{sec:cross-task} reconnects these task-specific views into a cross-task narrative over CNN-based, Transformer-based, foundation-model, and label-efficient paradigms. \textbf{Second, a shared cross-task analysis.} Each sub-task closes with \textbf{Common Problem} and \textbf{Comparison} paragraphs that map methods onto trade-offs, spanning motion-prior fidelity, accuracy--latency balance, supervision regime, identity stability, and long-tail robustness. \secref{sec:cross-task} then folds the shared issues into four dimensions: Temporal Consistency, Identity Preservation, Long-tail/Open-world Recognition, and Label Efficiency, all under the Annotation--Capacity--Latency Trilemma. It also catalogues failure modes, turning the survey from a method list into a diagnostic framework. \textbf{Third, temporal evidence.} \secref{sec:method} and \secref{sec:experiments} combine static benchmark tables with four performance-evolution figures on Cityscapes~\cite{cordts2016cityscapes}, VSPW~\cite{miao2021vspw}, YouTube-VIS~\cite{Yang2019VideoIS}, and KITTI-MOTS~\cite{voigtlaender2019mots}, giving a reproducible view of performance trends across tasks.}

{\figref{tab:structure} lays out the organization of this survey. It runs from problem definition and background through methods, datasets, and evaluation to future directions, keeping the five VSP tasks aligned across methods, resources, benchmarks, and cross-task discussion. The final outlook ties open-world, unified, multimodal, efficient, and LLM-based VSP to the failure modes and trends analyzed earlier.}

\section{Background}

\begin{figure}[!t]
\centering
\setlength{\tabcolsep}{1.mm}
\renewcommand{\arraystretch}{1}
\resizebox{0.82\linewidth}{!}{%
\begin{tabular}{c}
  \begin{tabular}{cccc}
    \includegraphics[width=0.18\textwidth]{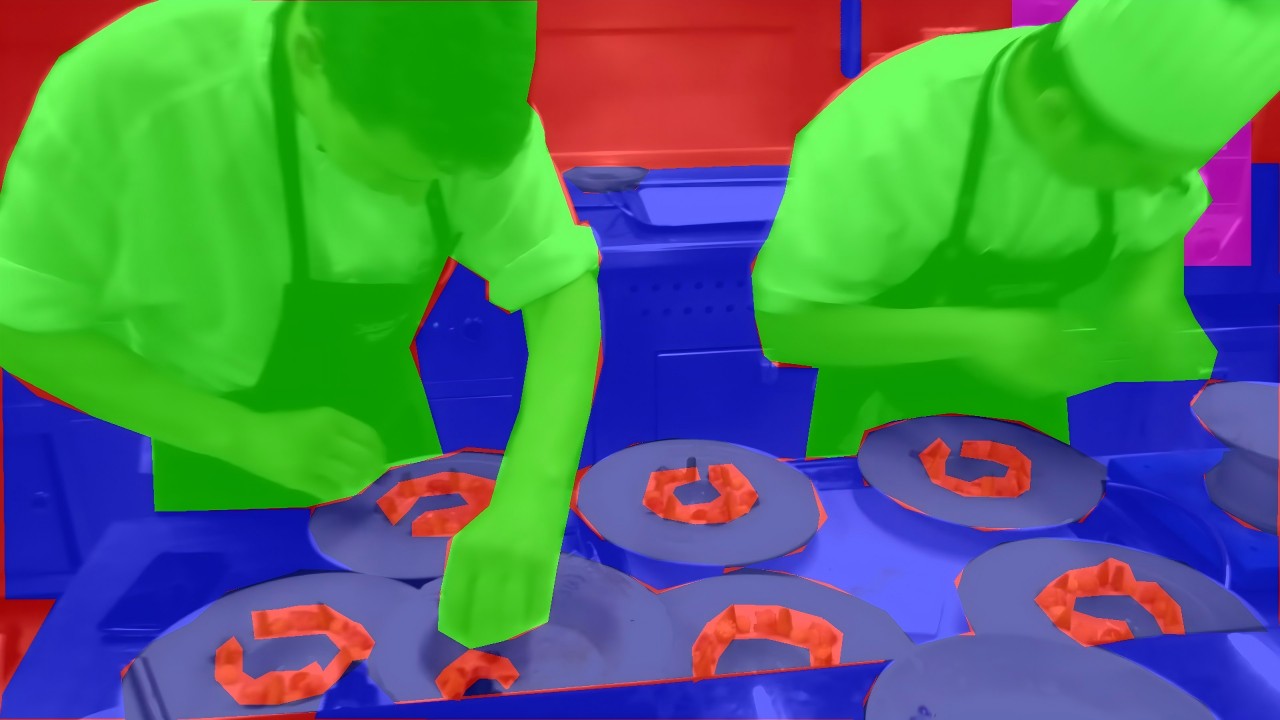} &
    \includegraphics[width=0.18\textwidth]{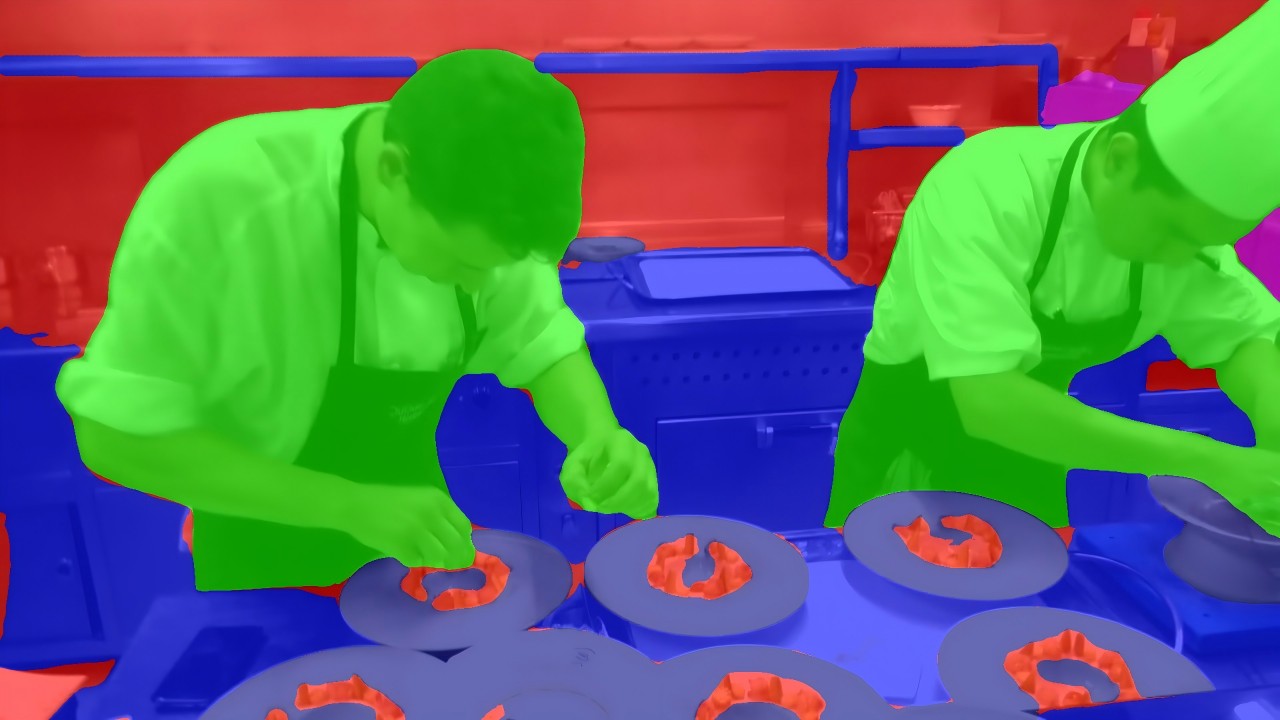} &
    \includegraphics[width=0.18\textwidth]{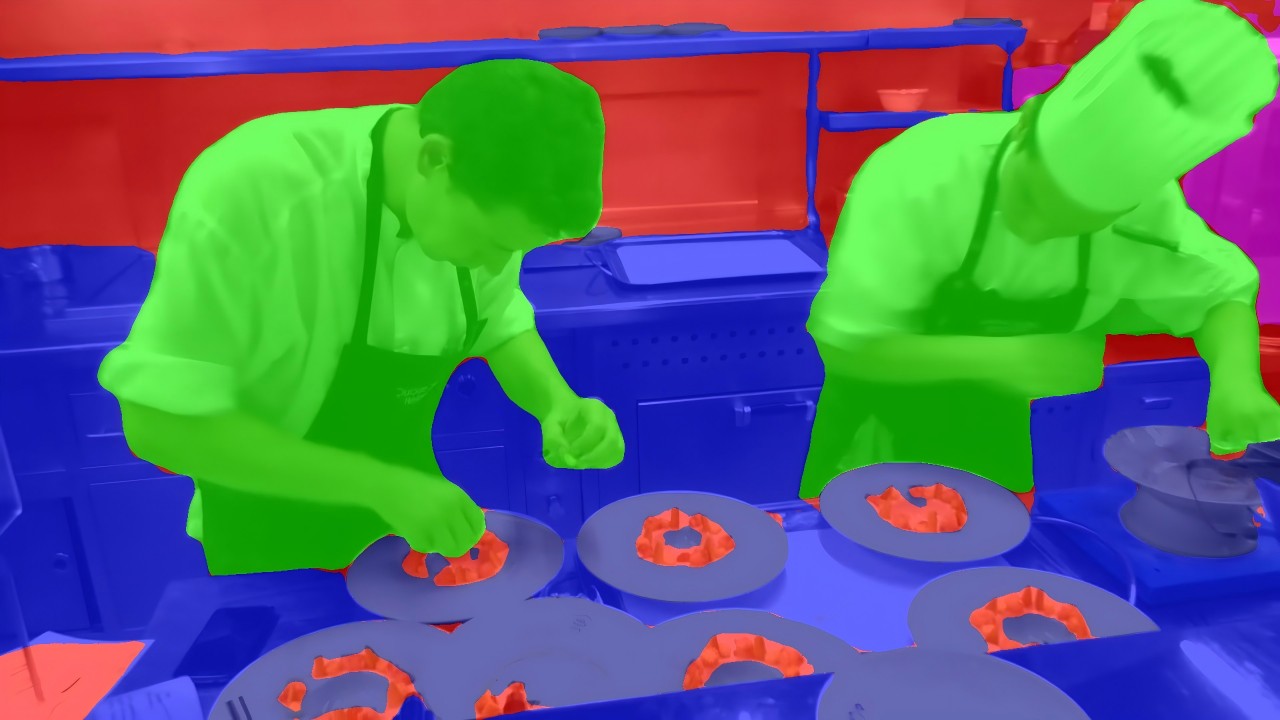} &
    \includegraphics[width=0.18\textwidth]{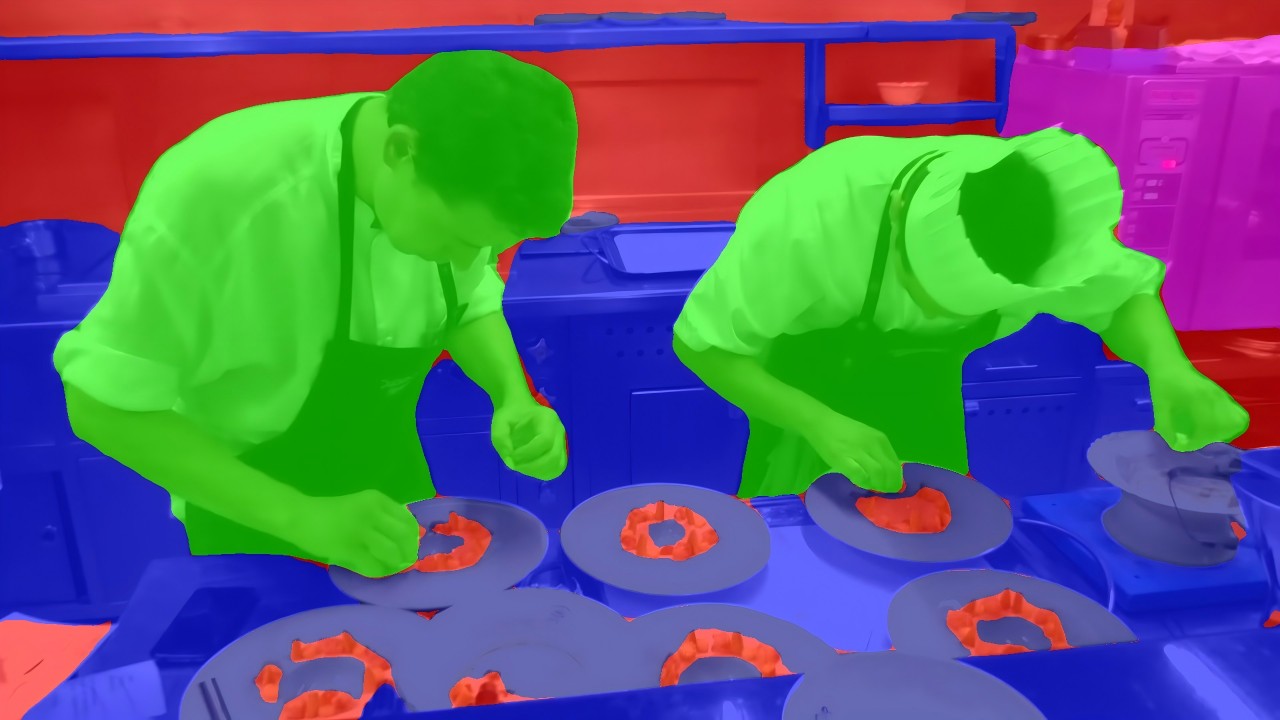} \\
    \multicolumn{4}{c}{\scriptsize (a) VSS}
  \end{tabular}\\
  \begin{tabular}{cccc}
    \includegraphics[width=0.18\textwidth]{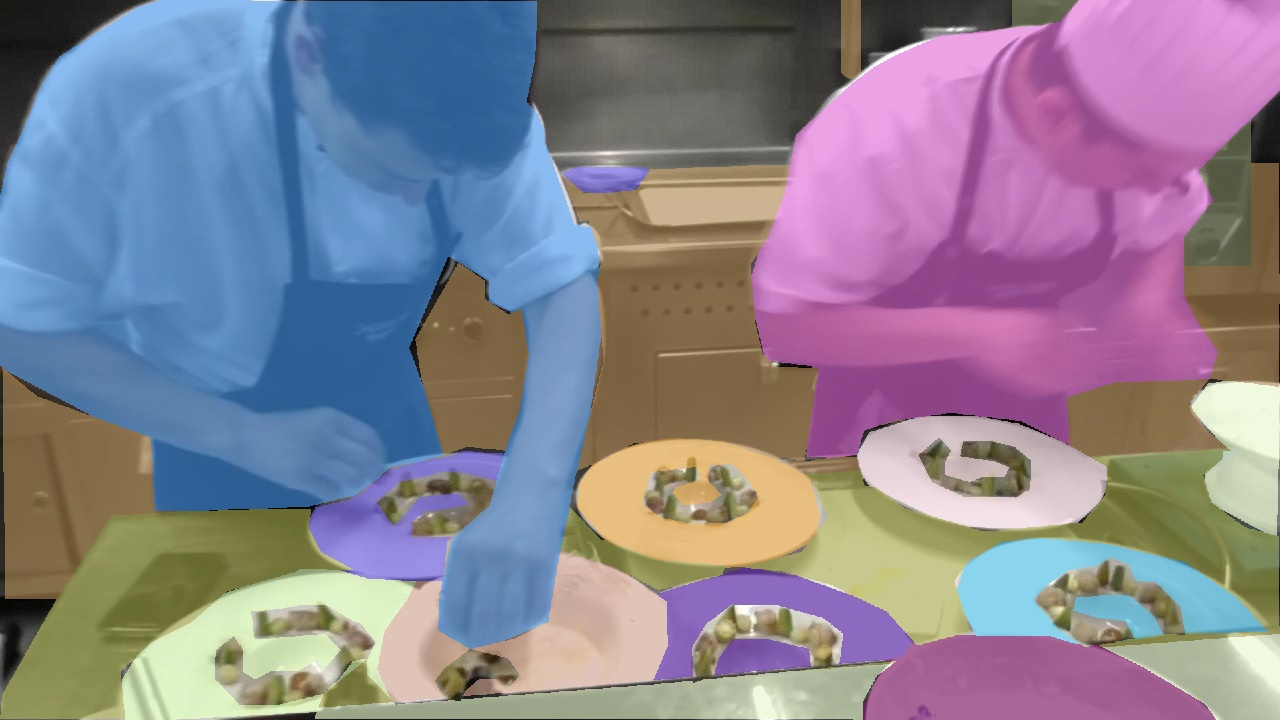} &
    \includegraphics[width=0.18\textwidth]{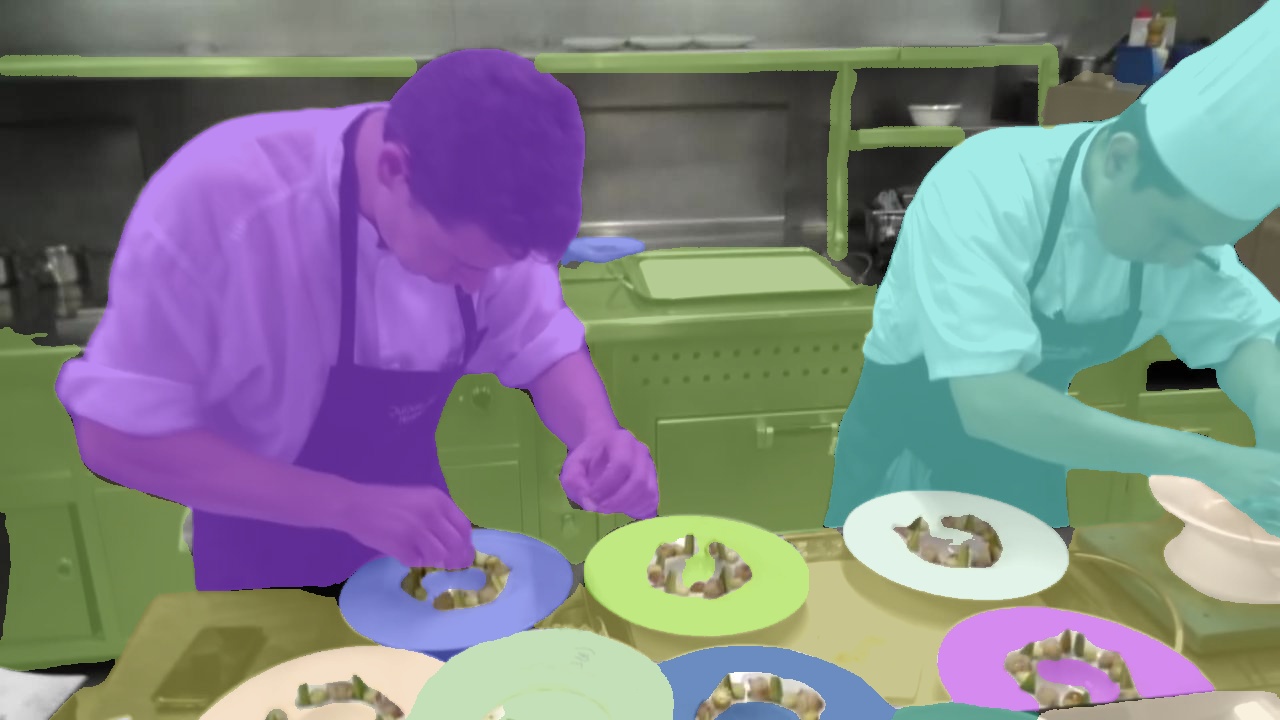} &
    \includegraphics[width=0.18\textwidth]{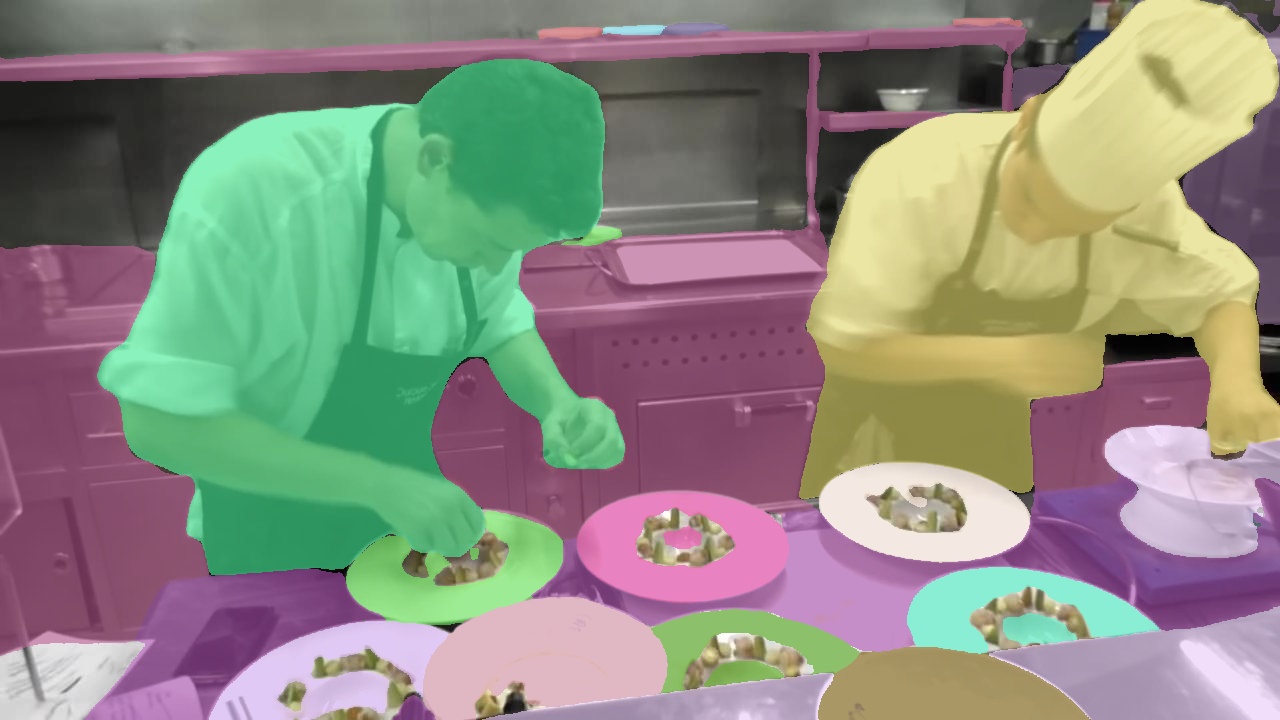} &
    \includegraphics[width=0.18\textwidth]{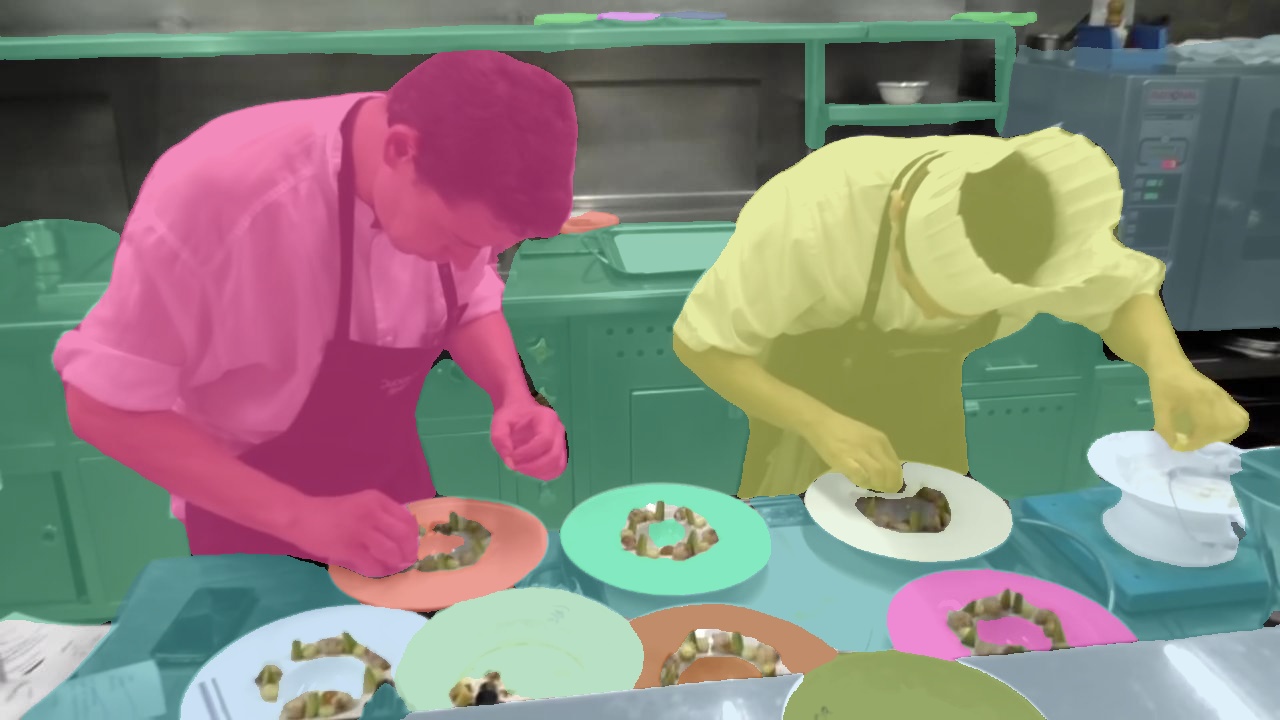} \\
    \multicolumn{4}{c}{\scriptsize (b) VIS}
  \end{tabular}\\
  \begin{tabular}{cccc}
    \includegraphics[width=0.18\textwidth]{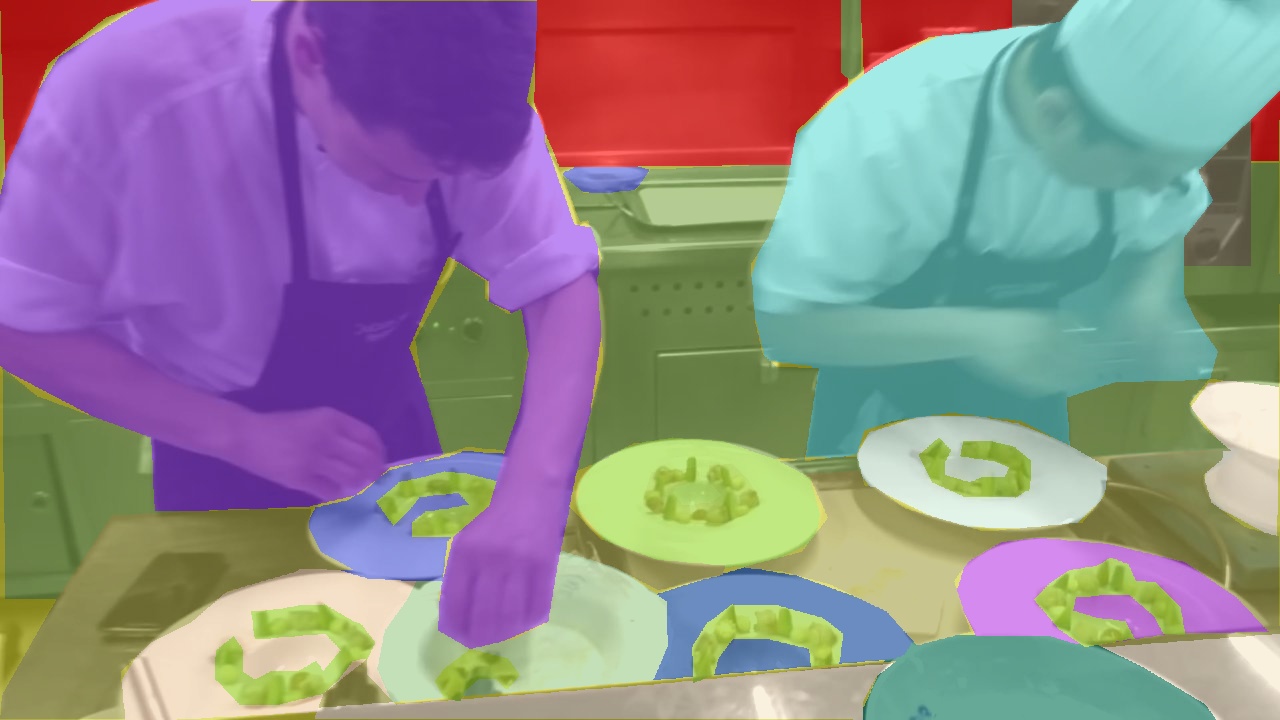} &
    \includegraphics[width=0.18\textwidth]{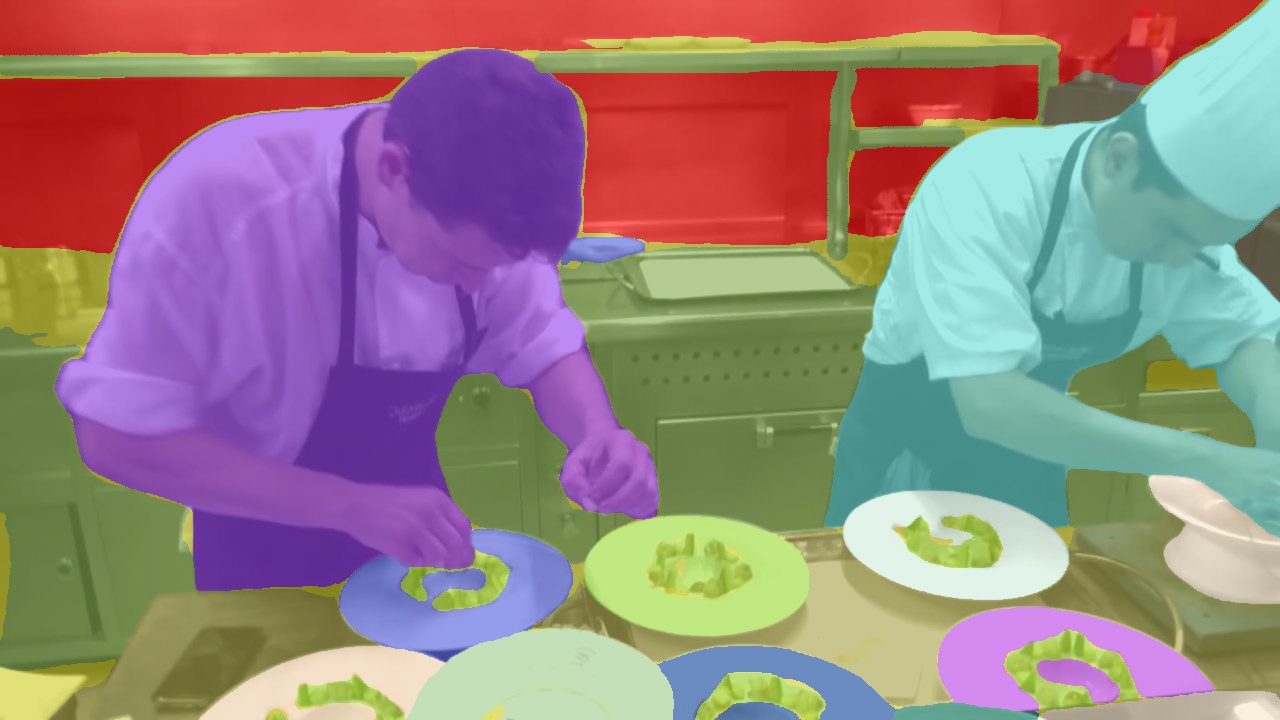} &
    \includegraphics[width=0.18\textwidth]{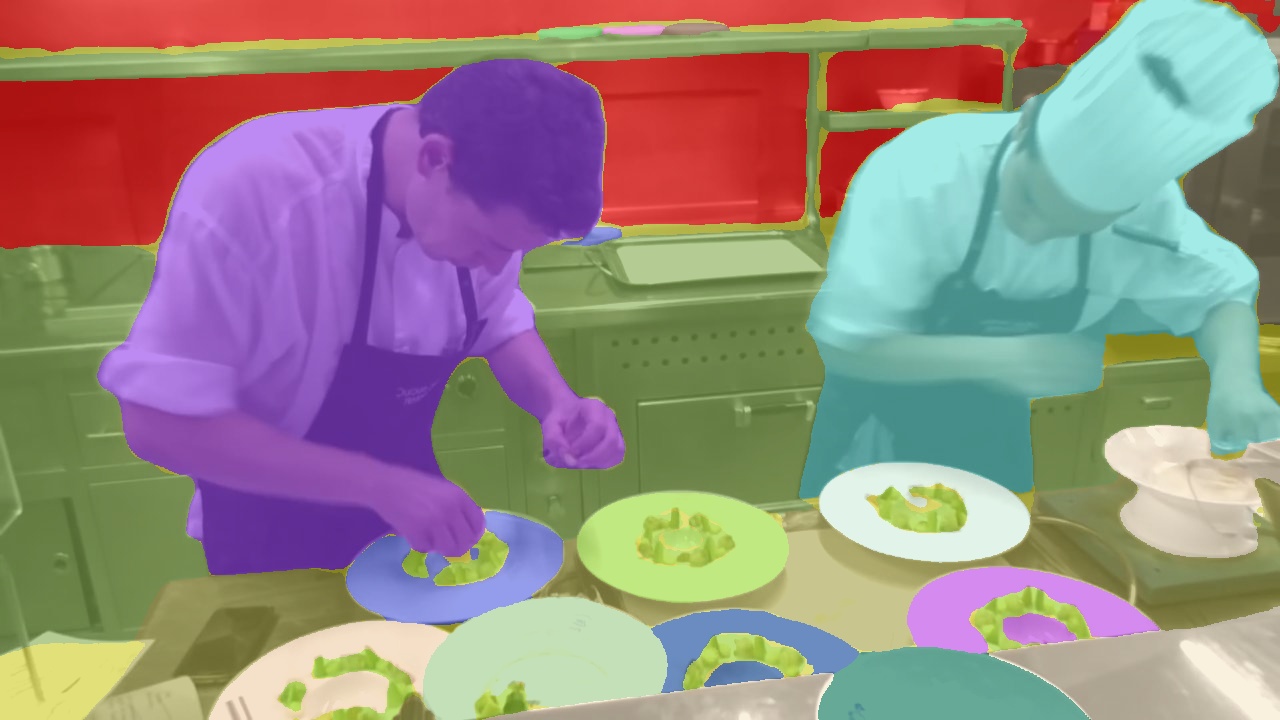} &
    \includegraphics[width=0.18\textwidth]{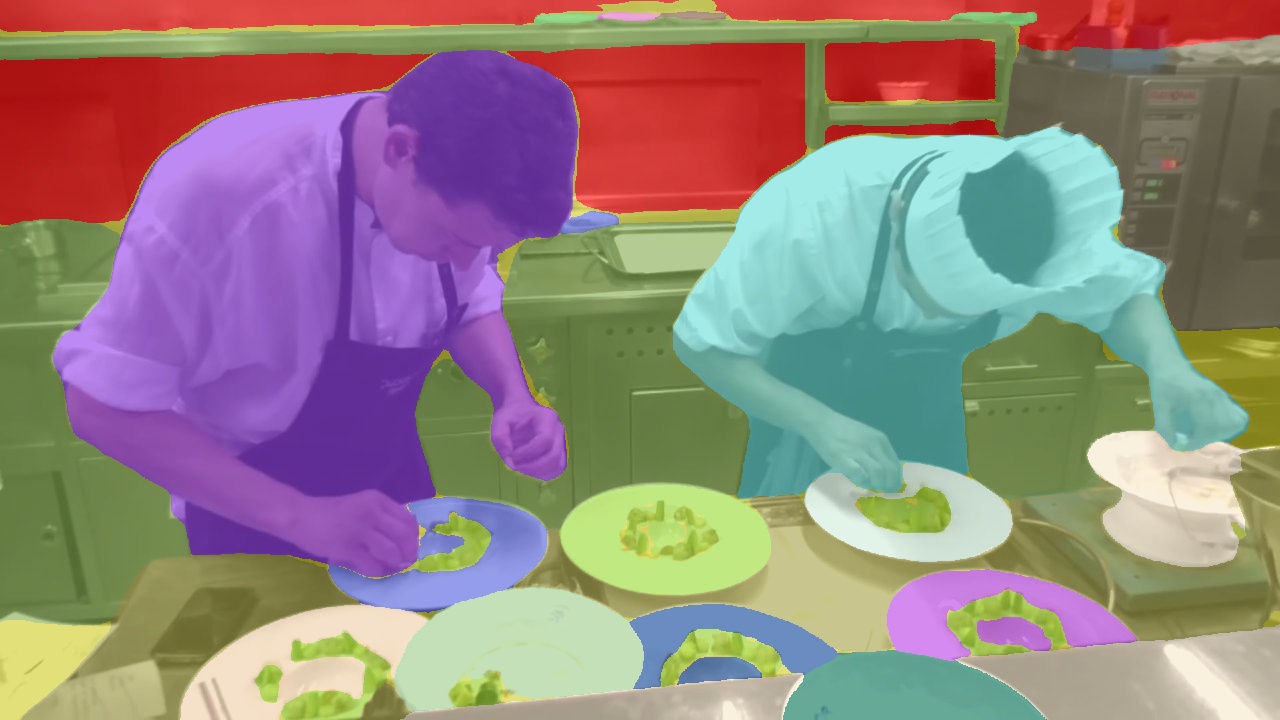} \\
    \multicolumn{4}{c}{\scriptsize (c) VPS}
  \end{tabular}\\
  \begin{tabular}{cccc}
    \includegraphics[width=0.18\textwidth]{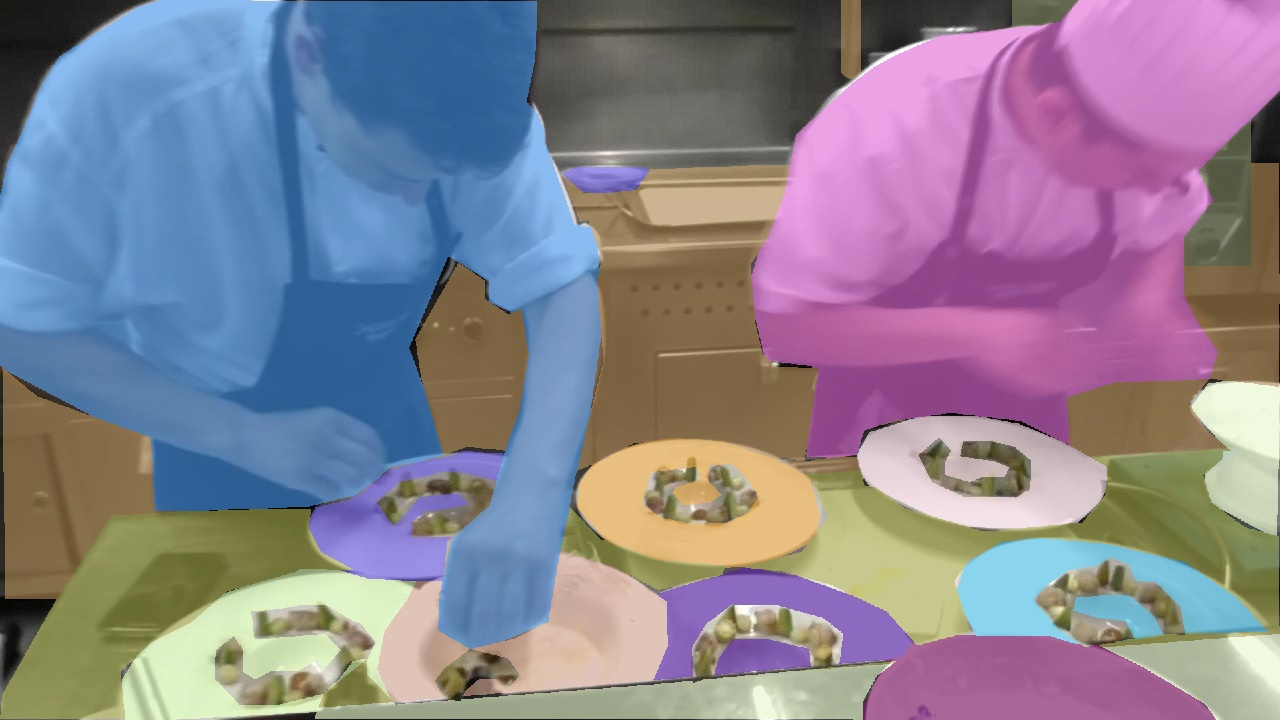} &
    \includegraphics[width=0.18\textwidth]{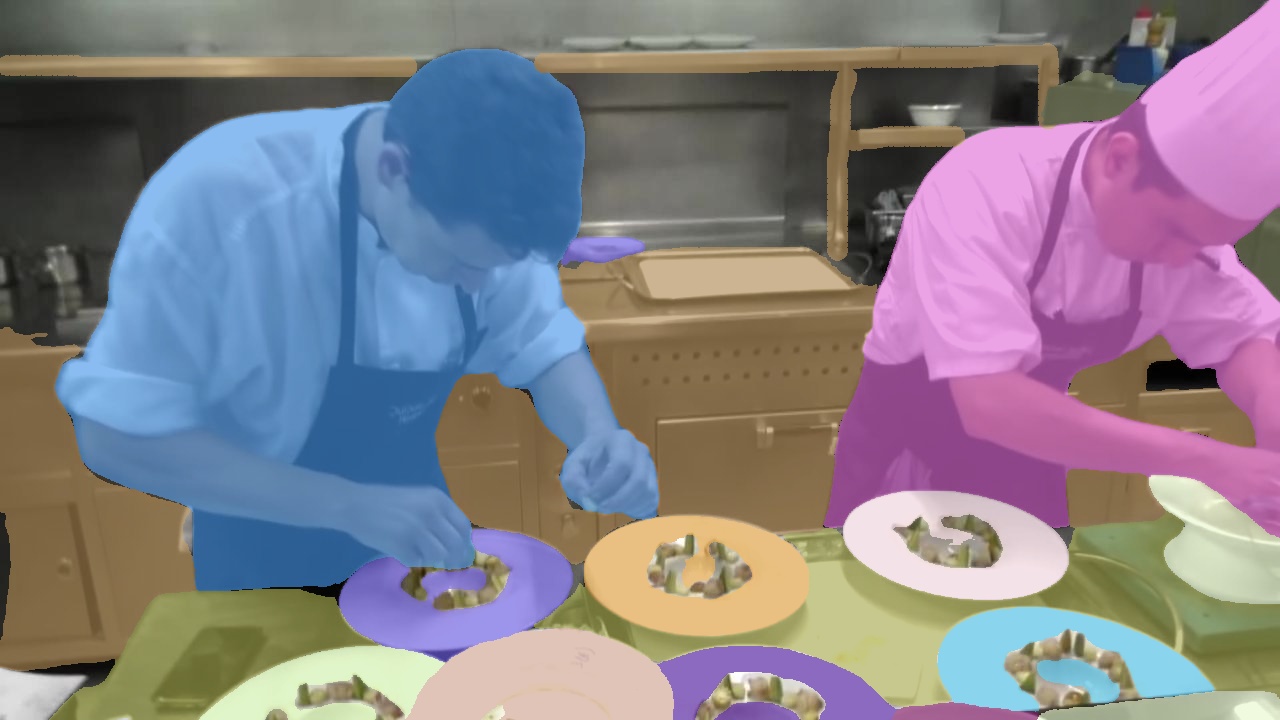} &
    \includegraphics[width=0.18\textwidth]{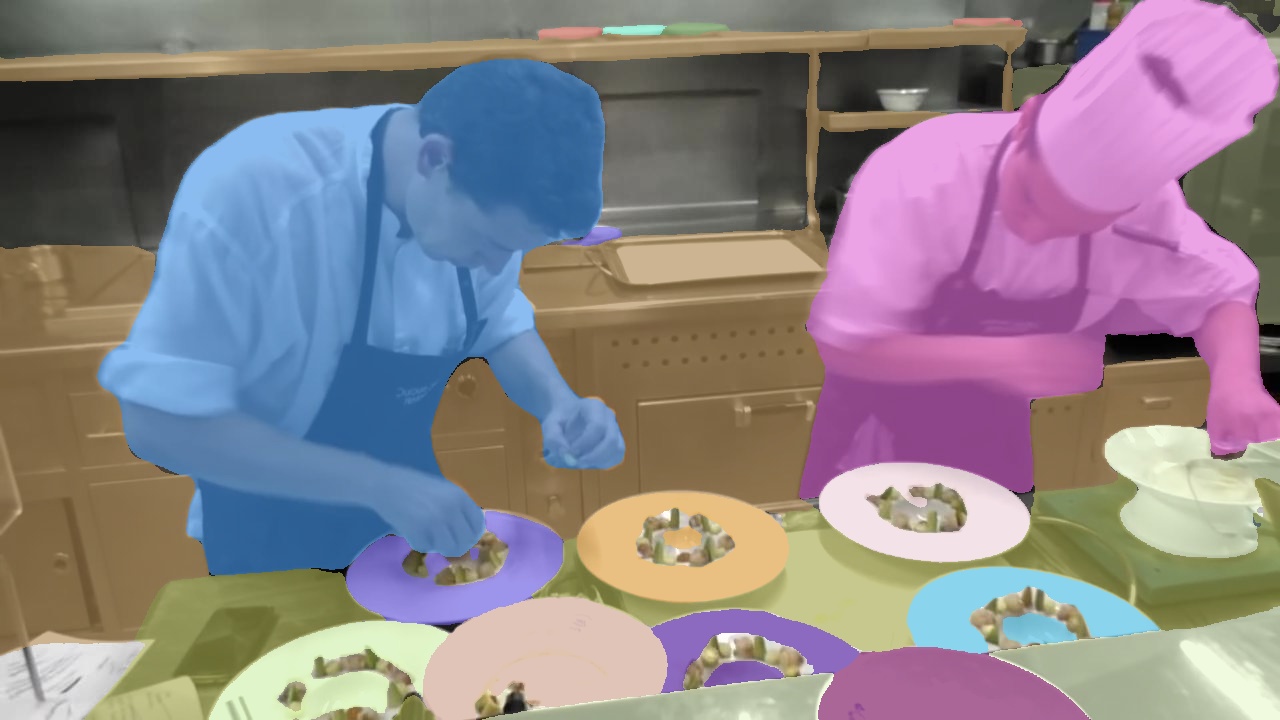} &
    \includegraphics[width=0.18\textwidth]{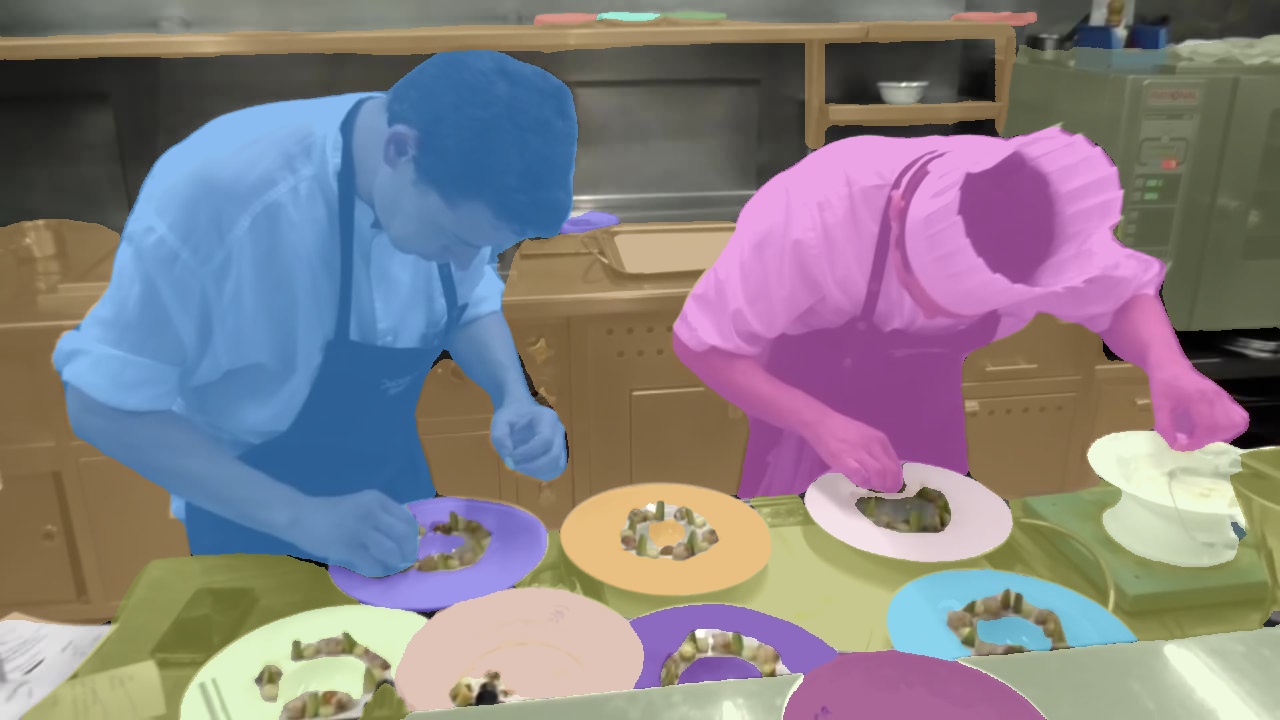} \\
    \multicolumn{4}{c}{\scriptsize (d) VTS}
  \end{tabular}\\
  \begin{tabular}{cccc}
    \includegraphics[width=0.18\textwidth]{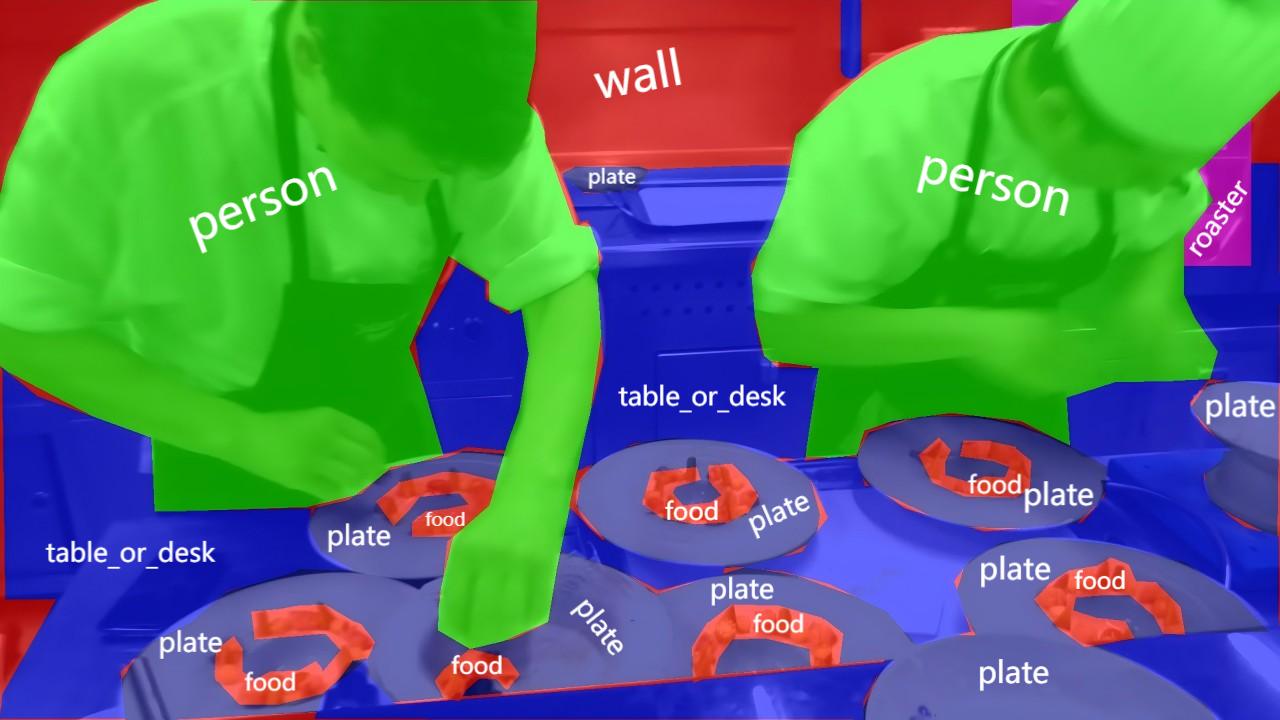} &
    \includegraphics[width=0.18\textwidth]{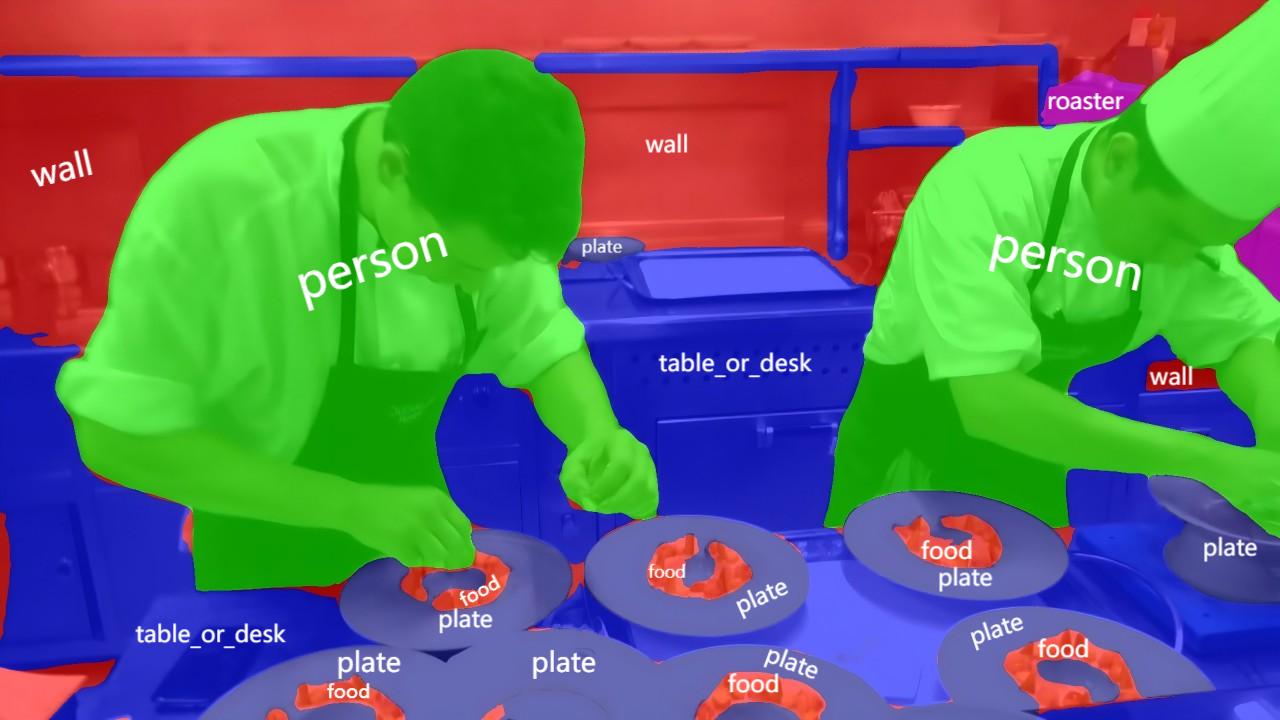} &
    \includegraphics[width=0.18\textwidth]{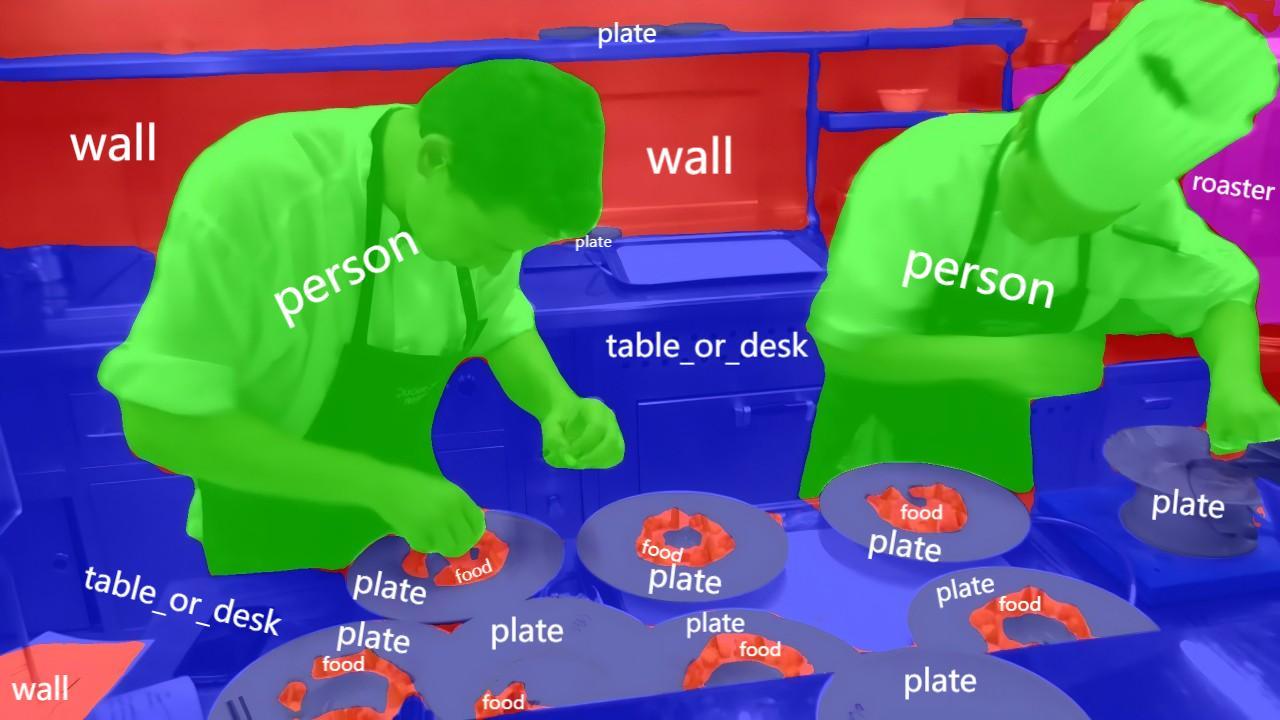} &
    \includegraphics[width=0.18\textwidth]{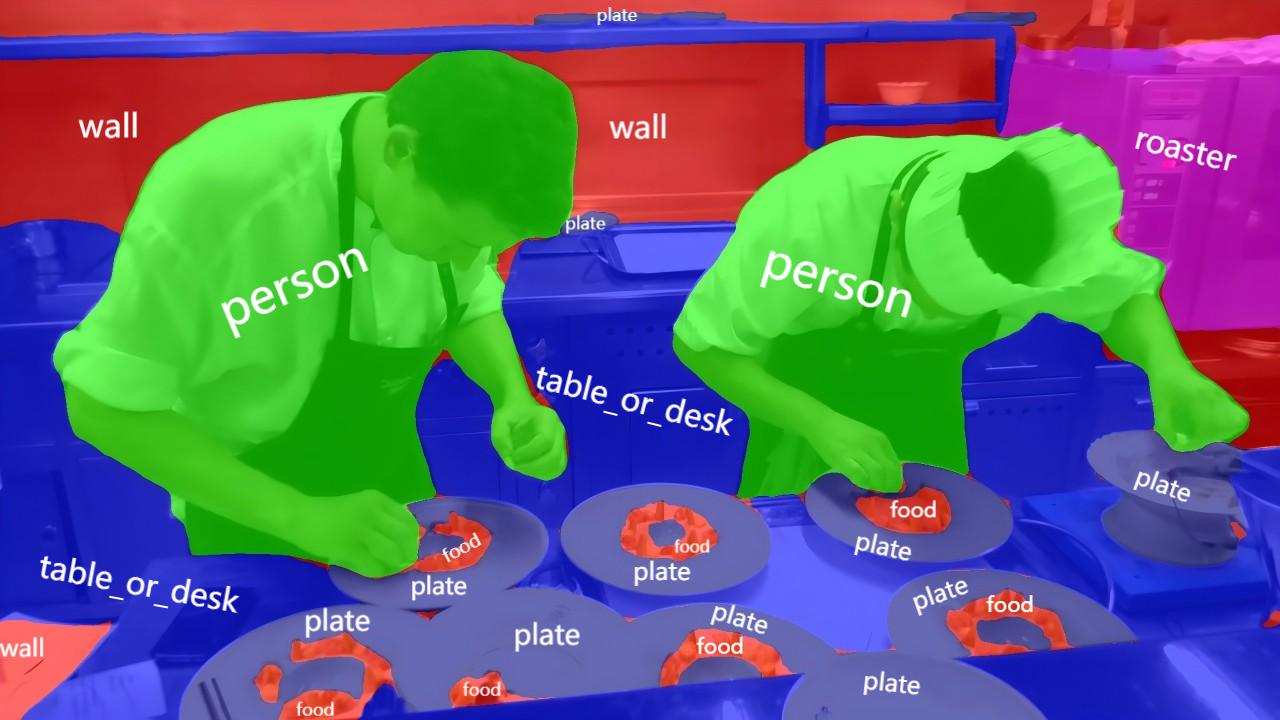} \\
    \multicolumn{4}{c}{\scriptsize (e) OVVS}
  \end{tabular}
\end{tabular}}
\vspace{-5pt}
\caption{{\textbf{Illustration of VSP tasks} (samples from VIPSeg~\cite{Miao2022LargescaleVP}). VSS: same color = same semantic class (no instance distinction). VIS/VPS: distinct colors per instance (VPS adds background semantics). VTS: instance color persists across frames. OVVS: novel categories beyond the predefined label set, with category names overlaid on each segment.}}
\label{fig:difference}
\vspace{-10pt}
\end{figure}

\subsection{Task Definition}\label{sec:def}
The five VSP tasks share one skeleton and differ only in what the output label must carry. Let \( V=\{I_t\}_{t=1}^{T} \) and \( Y \) denote the input video with \(T\) frames and the output segmentation space, respectively. A deep-learning VSP model then seeks an optimal mapping \( f^* : V \to Y \) from frames to segmentation labels, and \figref{fig:difference} visualizes VSS, VIS, VPS, VTS and OVVS with \( T=4 \) frames.

\para{Video Semantic Segmentation (VSS).} VSS is the base case, a per-frame class map that must stay stable across time. Given a clip \( V = \{I_t\}_{t=1}^{T} \in \mathbb{R}^{T \times H \times W \times 3} \) of \( T \) RGB frames, VSS seeks a mapping \( f^{\mathrm{VSS}} \) that returns a pixel-wise class map \( Y_t \) per frame under a cross-frame consistency constraint,
\begin{equation}
f^{\mathrm{VSS}}: \; V \;\longmapsto\; Y^{\mathrm{VSS}} = \{Y_t\}_{t=1}^{T}, \qquad Y_t \in \mathcal{C}^{H \times W}, \qquad Y_t \;\approx\; \mathcal{T}_{t-1\to t}(Y_{t-1}),
\end{equation}

where \( \mathcal{C} \) is a fixed closed label set (\eg, 19 Cityscapes classes), \( \mathcal{T}_{t-1\to t}(\cdot) \) propagates the previous prediction via flow or learned correspondences, and ``\( \approx \)'' encodes temporal consistency (\eg, evaluated by \( \mathrm{mVC}_n \)). This approximate equality separates VSS from per-frame image semantic segmentation, and the absence of instance identity in \( Y_t \) separates it from VIS below.

\para{Video Instance Segmentation (VIS).} VIS adds an explicit instance-identity dimension on top of VSS, and outputs \( N \) instance \emph{tubes} that span the clip,
\begin{equation}
f^{\mathrm{VIS}}: \; V \;\longmapsto\; Y^{\mathrm{VIS}} = \bigl\{\,(M_i,\, c_i,\, \mathrm{id}_i)\,\bigr\}_{i=1}^{N}, \qquad M_i \in \{0,1\}^{T \times H \times W},
\end{equation}
where \( M_i \) is the spatio-temporal binary mask (the ``tube'') of the \( i \)-th instance, \( c_i \in \mathcal{C} \) its class, and \( \mathrm{id}_i \) a unique identifier valid over the \( T \)-frame horizon. The temporal constraint therefore concerns identity consistency rather than class consistency alone: an adjacent-frame change in \( \mathrm{id}_i \), a missed instance, or an incorrect class assignment can create an identity error or a segmentation error, depending on which component changes.

\para{Video Panoptic Segmentation (VPS).} VPS fuses the previous two, merging the per-pixel class map of VSS with the instance tubes of VIS into a single non-overlapping partition, distinguishing \( K \) \emph{stuff} regions (uncountable semantics, \eg, road, sky) from \( N \) \emph{thing} tubes (countable objects),
\begin{equation}
f^{\mathrm{VPS}}: \; V \;\longmapsto\; Y^{\mathrm{VPS}} = \underbrace{\{(R_k,\, c_k^{\mathrm{stuff}})\}_{k=1}^{K}}_{\text{stuff regions, no identity}} \;\cup\; \underbrace{\{(M_i,\, c_i^{\mathrm{thing}},\, \mathrm{id}_i)\}_{i=1}^{N}}_{\text{thing tubes, identity-bearing}},
\end{equation}
where \( R_k \subset \{1,\ldots,T\}\!\times\!\{1,\ldots,H\}\!\times\!\{1,\ldots,W\} \) is a semantically continuous stuff region and the thing branch reuses the VIS tube formulation. The two branches are mutually exclusive at every spatio-temporal location, so every voxel receives exactly one (class, identity) pair and the canonical VPQ metric aggregates tube-level IoU over both. The extra requirement of cross-frame stuff continuity, jointly evaluated with thing identity, makes VPS strictly harder than either VSS or VIS alone.

\para{Video Tracking \& Segmentation (VTS).} VTS inherits the VIS tube output \( \mathcal{Y}^{\mathrm{VTS}} = \{(M_i, c_i, \mathrm{id}_i)\}_{i=1}^{N} \) but assumes a horizon \( T \) far longer than the few-second clips on which VIS is evaluated, shifting the focus from per-frame segmentation to long-horizon identity preservation,
\begin{equation}
f^{\mathrm{VTS}}: \; V \;\longmapsto\; Y^{\mathrm{VTS}}, \qquad \mathrm{id}_i^{(t)} = \mathrm{id}_i^{(t')} \;\;\forall\, (t, t') \in \{1,\ldots,T\}^{2} \;\;\text{with}\;\; |t - t'| \to \infty,
\end{equation}
where identity must persist even across occlusions and re-entries into the field of view. Evaluation therefore migrates from instantaneous AP to cumulative MOTSA, sMOTSA and IDS, and the signature failure modes become appearance drift, large camera motion and occlusion--reappearance.

\para{Open-Vocabulary Video Segmentation (OVVS).} OVVS departs from the four tasks above not in time but in label space: the closed set \( \mathcal{C} \) is replaced by an open semantic manifold spanned by a text encoder \( \phi(\cdot) \), so for any test-time vocabulary \( \mathcal{V} \) (natural-language nouns possibly unseen during training), class assignment is decided by vision--language similarity,
\begin{equation}
f^{\mathrm{OVVS}}: \; (V,\, \mathcal{V}) \;\longmapsto\; \bigl\{\,(M_i,\, c_i^{(t)},\, \mathrm{id}_i)\,\bigr\}_{i=1}^{N}, \qquad c_i^{(t)} = \arg\max_{v \in \mathcal{V}}\,\bigl\langle\,\psi\!\left(M_i^{(t)} \odot I_t\right),\; \phi(v)\,\bigr\rangle,
\end{equation}
where \( \psi(\cdot) \) is the visual encoder co-located with the text-embedding space and \( \langle\cdot,\cdot\rangle \) denotes inner product. This single substitution turns the closed-set classifier of any host task (VSS/VIS/VPS/VTS) into a vision--language alignment, but introduces three video-specific difficulties: \emph{(i) granularity ambiguity}, the same pixel may match multiple vocabulary levels (\eg, ``vehicle''/``car''/``sedan'') that closed-vocabulary protocols hard-code at the dataset level; \emph{(ii) temporal alignment drift}, since \( \phi \) and \( \psi \) are typically pre-trained on still images, the same object across frames can project to noticeably different textual neighborhoods under pose, illumination or motion blur, producing a ``semantic flicker'' impossible in closed-vocabulary VSP; and \emph{(iii) coupled long-tail and zero-shot generalization}, accuracy on seen classes and coherent masks for unseen ones compete for representational capacity, explaining the persistent gap between OVVS and fully-supervised VSP.

\revise{\tabref{tab:task_comparison} condenses the five formulations above, contrasting the tasks by output structure, temporal constraint, label space, and primary challenge. Read column-wise, it makes the relative complexity explicit: the output structure grows from per-frame class maps to identity-bearing tubes and mixed thing/stuff partitions, the temporal constraint tightens from label consistency to long-horizon identity persistence, and the label space finally opens from a fixed set to an unbounded vocabulary.}

\begin{table*}[!t]
\centering
\setlength{\tabcolsep}{1.5mm}
\caption{\revise{\textbf{Comparison of the five VSP tasks} by output structure, temporal constraint, label space, and primary challenge (cf.\ the formulations in \secref{sec:def}).}}
\vspace{-10pt}
\label{tab:task_comparison}
\revise{\resizebox{0.95\linewidth}{!}{%
\begin{tabular}{l|l|l|l|l}
\hline
\textbf{Task} & \textbf{Output Structure} & \textbf{Temporal Constraint} & \textbf{Label Space} & \textbf{Primary Challenge} \\
\hline
VSS & per-frame class maps $\{Y_t\}_{t=1}^{T}$ (no identity) & cross-frame label consistency (mVC$_n$) & closed, thing $+$ stuff & temporal flicker \vs latency \\
VIS & thing tubes $\{(M_i, c_i, \mathrm{id}_i)\}$ & clip-level identity consistency & closed, thing only & identity switch under occlusion \\
VPS & stuff regions $\cup$ thing tubes (non-overlapping) & joint class $+$ identity consistency & closed, thing $+$ stuff & thing/stuff balance, long tail \\
VTS & thing tubes over long streams & long-horizon identity persistence & closed, thing only & occlusion--reappearance re-ID \\
OVVS & host-task output with text-matched classes & host constraint $+$ stable vision--language alignment & open vocabulary $\mathcal{V}$ & granularity ambiguity, semantic flicker, zero-shot generalization \\
\hline
\end{tabular}}}
\vspace{-10pt}
\end{table*}

\subsection{History}\label{sec:dm}

Image segmentation research began with early boundary-detection methods~\cite{do1961machine}, many of which later migrated to video. The first VSP attempts grouped pixels by intensity similarity~\cite{xu2012evaluation,grundmann2010efficient,chang2013video,ochs2013segmentation,papazoglou2013fast}, without explicit spatio-temporal modeling. A second wave injected optical-flow~\cite{narayana2013coherent,sun2012layered,bai2010dynamic,liu2014steadyflow}, graph~\cite{grundmann2010efficient,khoreva2015classifier} and graph-cut~\cite{peng2014jf} priors, chasing the temporal stability the first wave missed. The first real inflection point arrived with deep CNNs, which swapped hand-crafted motion and appearance cues for end-to-end FCNs~\cite{shelhamer2017fully,chen2018deeplab,zhao2017pyramid,ding2018context}. This crystallized the ``per-frame FCN $+$ post-hoc smoothing'' template, trading interpretable priors for learned hierarchical features but inheriting per-frame flicker, which flow~\cite{jin2017video} or CRF~\cite{yuan2022neural} reasoning then patched after the fact. Three further inflection points, each spawning its own branch, grew out of this CNN template. The second inflection added explicit instance identity~\cite{Yang2019VideoIS}, exposing association and occlusion as first-order problems rather than post-processing artefacts. The third shifted the binding constraint from accuracy to annotation and latency, motivating label-efficient self-/unsupervised learning~\cite{Chen2020NaiveStudentLS,Hoyer2020ThreeWT,Lao2023SimultaneouslySA} and real-time architectures~\cite{shelhamer2016clockwork,paul2020efficient,park2022real,xu2018dynamic,li2018low}. The fourth replaced hand-designed temporal fusion with Transformer-based attention and query-based set prediction~\cite{Sun2022CoarsetoFineFM,sun2022mining}, unifying detection, segmentation and association at the cost of quadratic compute and data-hungry training.

\begin{figure*}[!t]
\centering
\includegraphics[width=0.92\linewidth]{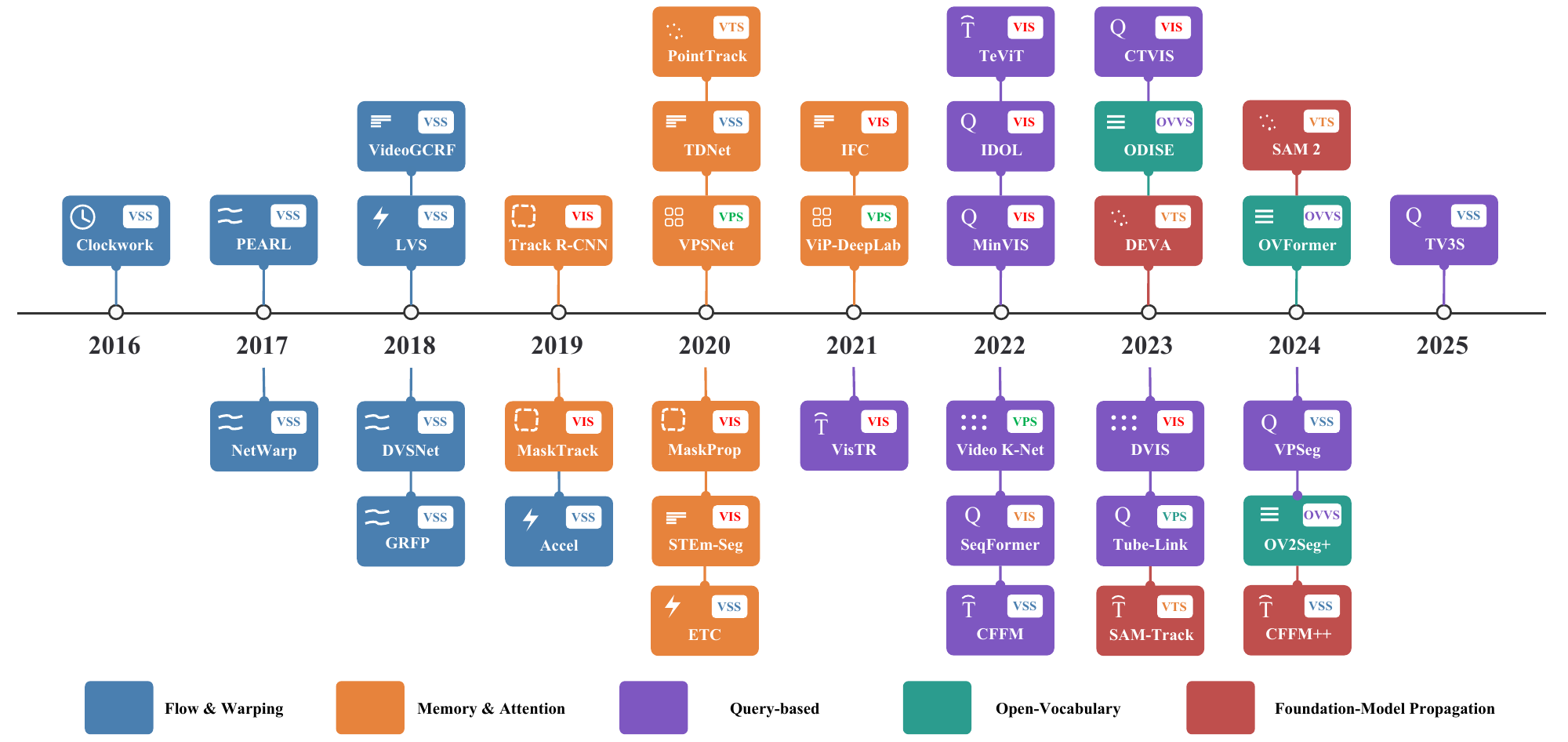}
\caption{\revise{\textbf{Architectural evolution of Video Scene Parsing (2016--2025).}
Representative methods are organized into five paradigm bands and
color-coded by target task (VSS/VIS/VPS/VTS/OVVS); all methods also appear
in the per-task summary tables in \secref{sec:method}.}}
\label{fig:vsp_evolution}
\vspace{-23pt}
\end{figure*}

We make this architectural evolution explicit in \figref{fig:vsp_evolution}, laid out as five chronological paradigms separated by the four inflection points above. Each shift resolved the dominant limitation of its predecessor while introducing a new one, a pattern the cross-task analysis in \secref{sec:cross-task} makes explicit.

\subsection{Related Research Areas}\label{sec:cd}

\revise{VSP does not sit in isolation, but it borrows heavily from three neighboring fields, each solving a narrower slice of the same underlying puzzle, with image segmentation supplying the per-pixel segmentation concept, video object segmentation providing the temporal machinery, and video object detection supplementing with object-centric enhancements. What follows traces each contribution in turn and marks exactly where VSP breaks away.}

\para{Image Semantic Segmentation.} \revise{Image semantic segmentation has driven VSP's progress from the start~\cite{luo2017deep,ahn2018learning,zhao2018icnet,wang2019panet,araslanov2020single,fan2020cian,xie2026make}. Early VSS approaches~\cite{mahasseni2017budget,siam2017convolutional} simply applied image segmenters frame by frame. Later work learned to exploit spatio-temporal consistency for both accuracy and efficiency, yet image segmentation remains its methodological foundation. This dependence runs deep, since a plain ViT with no segmentation-specific decoder can already act as a competitive image segmenter~\cite{kerssies2025your}. VidEoMT~\cite{norouzi2026videomt} then carries the idea into video, showing that the backbone-as-segmenter principle holds across the image and video boundary relevant to VSP.}

\noindent \revise{\emph{Distinction from VSP.} The two diverge along three axes: \textit{(i) temporal constraints}, image segmentation needs only per-frame pixel correctness, VSP additionally enforces temporal consistency (mVC, clip-AP, VPQ, IDS and their zero-shot variants); \textit{(ii) output space}, from $H\!\times\!W$ to a $T\!\times\!H\!\times\!W$ space-time volume, with VIS/VPS/VTS further preserving cross-frame identity; \textit{(iii) failure modes}, image segmenters fail at category boundaries, VSP additionally at instance birth/death, identity switches and temporally fragmented rare-class predictions.}

\para{Video Object Segmentation.} \revise{Add a temporal dimension to the segmentation problem and the result is VSP's closest cousin, Video Object Segmentation (VOS). VOS provides VSP with many of its spatio-temporal building blocks, and splits into three lineages, each with a clear VSP imprint. \textit{(i) Online fine-tuning} (OSVOS~\cite{caelles2017one}, PReMVOS~\cite{luiten2018premvos}) adapts a generic mask network to a specific instance from the first-frame annotation, and PReMVOS pushed this further by fusing proposal generation, refinement and ReID-based merging into the now-standard ``segment-then-track'' pipeline that VIS and VTS inherited. \textit{(ii) Mask propagation and matching} (MaskTrack~\cite{perazzi2017learning}, RVOS~\cite{ventura2019rvos}) cast segmentation as cross-frame warping or recurrent prediction, showing that dense masks can propagate without per-frame re-detection. \textit{(iii) Spatio-temporal memory networks} (STM~\cite{oh2019video}, Episodic Graph Memory~\cite{lu2020video}, Global Context VOS~\cite{li2020fast}, ERMN~\cite{xie2021efficient}, Collaborative VOS~\cite{yang2020collaborative}, STCN~\cite{cheng2021rethinking}, XMem~\cite{cheng2022xmem}, Cutie~\cite{cheng2024putting}) recast temporal modeling as read and write operations over an external feature bank, where XMem's Atkinson--Shiffrin three-tier memory extends VOS to minute-scale videos at constant memory.}

\noindent \revise{\emph{Distinction from VSP.} Despite shared temporal machinery, VOS and VSP diverge along three axes: \textit{(i) problem framing}, VOS is class-agnostic and instance-bounded, a first-frame mask (semi-supervised) or saliency cue (unsupervised) specifies the target so categories never enter, whereas VSP classifies every pixel (VSS), discovers/labels every instance (VIS), unifies thing/stuff (VPS) or preserves trajectory identity (VTS); \textit{(ii) supervision regime}, VOS (DAVIS, YouTube-VOS) is one-/few-shot, VSP (Cityscapes-VSS, VSPW, YouTube-VIS, VIPSeg, KITTI-MOTS) is dense fully-supervised; \textit{(iii) failure modes}, VOS stresses appearance drift and re-ID of a known target, VSP additionally faces class confusion, instance birth/death and stuff flicker. VOS thus feeds VSP architecturally (memory read/write, query-key matching, mask propagation), not semantically.}

\para{Video Object Detection.} {Where VOS follows one specified target through time, Video Object Detection (VOD) broadens the question to localizing and classifying every object of interest, frame after frame. These detectors inject temporal cues into image-level detection pipelines~\cite{zhu2018towards,zhu2017flow,wang2018fully,sun2021mamba,zhang2021dynamic,meeran2024sam}, so they share with instance-level VSP the core challenges of temporal consistency, motion blur and occlusion. That shared machinery also lays useful groundwork for VSP, where scene dynamics and object interactions are central.}

\noindent \revise{\emph{Distinction from VSP.} VOD and VIS share temporal building blocks (feature aggregation, cross-frame association, motion compensation), yet diverge along three axes: \textit{(i) geometric granularity}, VOD predicts a bbox (4 scalars), VSP a pixel mask ($H\!\times\!W$), an order-of-magnitude tighter bar on boundary accuracy, thin structures and occlusion; \textit{(ii) semantic coverage}, VOD handles only fixed-vocabulary \emph{thing} categories, whereas VSP covers dense semantics (VSS), thing+stuff under VPQ (VPS), thing-only dense masks (VIS/VTS) or open-vocabulary recognition (OVVS); \textit{(iii) identity preservation}, VOD associates boxes only, VTS additionally enforces temporally coherent \emph{mask shape}, ruling out the ``consistent boxes but flickering masks'' mode.}

\begin{table*}[!t]
\centering
\setlength{\tabcolsep}{1.5mm}
\caption{\textbf{Summary of essential characteristics for reviewed VSS methods.}}
\vspace{-10pt}
\label{table:VSS_methods_summary}
\resizebox{0.92\linewidth}{!}{%
\begin{tabular}{c|c|c|c|c}
\hline
\textbf{Methods} & \textbf{Publication} & \textbf{Core Architectures} & \textbf{Technical Features} & \textbf{Training Datasets} \\
\hline
Clockwork \cite{shelhamer2016clockwork} & ECCV-W'2016 & FCN-8s & Adaptive layer-wise clock scheduling & NYUDv2 \cite{silberman2012indoor}/Cityscapes \cite{cordts2016cityscapes}\\
NetWarp \cite{Gadde2017SemanticVC} & ICCV'2017 & FCN + Flow Warping & Optical-flow-based representation warping & CamVid \cite{brostow2008segmentation}/Cityscapes \cite{cordts2016cityscapes}\\
PEARL \cite{jin2017video} & ICCV'2017 & FCN + GAN & Predictive feature learning & CamVid \cite{brostow2008segmentation}/Cityscapes \cite{cordts2016cityscapes}\\
VideoGCRF \cite{Chandra2018DeepSR} & CVPR'2018 & FCN+CRF & Gaussian CRF & DAVIS \cite{perazzi2016benchmark,pont20172017}/CamVid \cite{brostow2008segmentation} \\
EUVS \cite{huang2018efficient} & ECCV'2018 & Bayesian CNN & Flow-guided feature aggregation & CamVid \cite{brostow2008segmentation}\\
LVS \cite{li2018low} & CVPR'2018 & FCN & Keyframe selection & CamVid \cite{brostow2008segmentation}/Cityscapes \cite{cordts2016cityscapes}\\
GRFP \cite{nilsson2018semantic} & CVPR'2018 & FCN + GRU & Temporal feature aggregation & CamVid \cite{brostow2008segmentation}/Cityscapes \cite{cordts2016cityscapes}\\
DVSNet \cite{xu2018dynamic} & CVPR'2018 & FCN+RL & Keyframe selection & Cityscapes \cite{cordts2016cityscapes}\\
Accel \cite{Jain2018AccelAC} & CVPR'2019 & FCN & Keyframe selection & KITTI \cite{geiger2012KITTI} \\
SSeg \cite{zhu2019improving} & CVPR'2019 & FCN & Weakly-supervised learning & CamVid \cite{brostow2008segmentation}/Cityscapes \cite{cordts2016cityscapes}\\
Naive-Student \cite{Chen2020NaiveStudentLS} & ECCV'2020 & FCN+KD & Semi-supervised learning & Cityscapes \cite{cordts2016cityscapes}\\
EFC \cite{Ding2019EveryFC} & AAAI'2020 & FCN & Temporal feature aggregation & CamVid \cite{brostow2008segmentation}/Cityscapes \cite{cordts2016cityscapes}\\
TDNet \cite{Hu2020TemporallyDN} & CVPR'2020 & Memory Network & Attention-based feature aggregation & CamVid \cite{brostow2008segmentation}/Cityscapes \cite{cordts2016cityscapes}/NYUDv2 \cite{silberman2012indoor}\\
ETC \cite{liu2020efficient} & ECCV'2020 & FCN + KD & Knowledge distillation & CamVid \cite{brostow2008segmentation}/Cityscapes \cite{cordts2016cityscapes}\\
DepthMix \cite{Hoyer2020ThreeWT} & CVPR'2021 & FCN & Semi-supervised learning & Cityscapes \cite{cordts2016cityscapes}\\
CFFM \cite{Sun2022CoarsetoFineFM} & CVPR'2022 & Transformer & Static and motional context learning & Cityscapes \cite{cordts2016cityscapes}/VSPW \cite{miao2021vspw}\\
MRCFA \cite{sun2022mining} & ECCV'2022 & Transformer & Attention to cross-frame affinities & Cityscapes \cite{cordts2016cityscapes}/VSPW \cite{miao2021vspw}\\
CIRKD \cite{Yang2022CrossImageRK} & CVPR'2022 & KD & Cross-image relation & CamVid \cite{brostow2008segmentation}/Cityscapes \cite{cordts2016cityscapes}/Pascal VOC \cite{everingham2015pascal}\\
IFR \cite{zhuang2022semi} & CVPR'2022 & FCN & Semi-supervised learning & CamVid \cite{brostow2008segmentation}/Cityscapes \cite{cordts2016cityscapes}\\
MVNet \cite{Ji2023MultispectralVS} & CVPR'2023 & Transformer & Multispectral input & MVSeg \cite{Ji2023MultispectralVS}\\
SSLTM \cite{Lao2023SimultaneouslySA} & CVPR'2023 & FCN + Transformer & Semi-supervised learning & Cityscapes \cite{cordts2016cityscapes}/VSPW \cite{miao2021vspw}\\
MPVSS \cite{weng2023mask} & NIPS'2023 & FCN + Transformer & Keyframe selection & Cityscapes \cite{cordts2016cityscapes}/VSPW \cite{miao2021vspw}\\
BLO \cite{yang2023pruning} & CVPR'2023 & Transformer & Keyframe selection & Cityscapes \cite{cordts2016cityscapes}/ADE20K \cite{zhou2019semantic}/Pascal VOC \cite{everingham2015pascal}\\
VPSeg \cite{Guo2024VanishingPointGuidedVS} & CVPR'2024 & Transformer & Vanishing point & ACDC \cite{sakaridis2021acdc}/Cityscapes \cite{cordts2016cityscapes}\\
TDC \cite{zhuang2024infer} & CVPR'2024 & FCN & Semi-supervised learning & CamVid \cite{brostow2008segmentation}/Cityscapes \cite{cordts2016cityscapes}\\
CFFM++ \cite{Sun2022LearningLA} & PAMI'2024 & Transformer & Global temporal context learning & CamVid \cite{brostow2008segmentation}/Cityscapes \cite{cordts2016cityscapes}\\
TV3S \cite{TV3S2025Hesham} & CVPR'2025 & Transformer + Mamba & State space & VSPW \cite{miao2021vspw}/Cityscapes \cite{cordts2016cityscapes}\\
\hline
\end{tabular}}
\vspace{-20pt}
\end{table*}

\section{Methods: A Survey}\label{sec:method}
\subsection{Video Semantic Segmentation}

{VSS extends image semantic segmentation to video, where naive per-frame inference ignores temporal continuity~\cite{jiao2021new}. We organize the methods chronologically and functionally into four directions: flow-based propagation for early temporal alignment, attention-based aggregation for learned cross-frame context, real-time designs for latency control, and semi-/weakly-supervised learning for reducing annotation cost.} \revise{\tabref{table:VSS_methods_summary} summarizes the essential characteristics of the reviewed VSS methods.}

\para{Flow-based Methods.} {Optical flow supplies a geometrically interpretable motion prior, estimating dense per-pixel displacement under brightness constancy for cross-frame transfer~\cite{huang2018efficient}. Work here splits into two routes. \textit{(i) Feature propagation} uses flow to align or adaptively update inter-frame features, as in NetWarp~\cite{Gadde2017SemanticVC}, PEARL~\cite{jin2017video} and DVSNet~\cite{xu2018dynamic}. \textit{(ii) Explicit spatio-temporal modeling} builds the motion reasoning into the model itself: GCRF~\cite{Chandra2018DeepSR} optimizes pixel consistency through deep spatio-temporal CRFs, STGRU~\cite{nilsson2018semantic} gates flow propagation under occlusion, Accel~\cite{Jain2018AccelAC} pairs low-resolution motion prediction with high-resolution refinement, EFC~\cite{Ding2019EveryFC} jointly optimizes flow and segmentation through bidirectional propagation, and ETC~\cite{liu2020efficient} couples per-frame inference with feature-level temporal aggregation. Both routes hit the same ceiling, since their accuracy is capped by the quality of the underlying flow.}

\para{Attention-based Methods.} {Attention mechanisms in VSS act primarily as spatio-temporal feature aggregators, and its development reads along three functional axes rather than as unrelated groups. \textit{(i) Temporal distillation and multi-scale fusion}: TDNet~\cite{Hu2020TemporallyDN} couples feature distillation with multi-scale aggregation through a lightweight temporal branch; CFFM~\cite{Sun2022CoarsetoFineFM} performs adaptive multi-scale fusion via a feature pyramid, and CFFM++~\cite{Sun2022LearningLA} extends it with global temporal-context extraction; MRCFA~\cite{sun2022mining} models complex motion through cross-frame affinity. \textit{(ii) Knowledge and cross-view transfer}: CIRKD~\cite{Yang2022CrossImageRK} distills temporal cues from low- to high-resolution features; MVNet~\cite{Ji2023MultispectralVS} fuses multi-spectral views; SSLTM~\cite{Lao2023SimultaneouslySA} jointly models short- and long-term temporal dependencies. \textit{(iii) Keyframe guidance and geometric priors}: MPVSS~\cite{weng2023mask} uses keyframe results to guide non-keyframe mask propagation, and VPSeg~\cite{Guo2024VanishingPointGuidedVS} injects vanishing-point priors for sparse-to-dense feature mining and motion fusion. The chronological trend is from lightweight temporal branches to richer global context and then to explicit geometric or keyframe priors.}

\para{Real-time Methods.} {Real-time VSS trims latency through three levers: lightweight architectures, adaptive scheduling and feature reuse. Clockwork~\cite{shelhamer2016clockwork} stages an FCN under stage-wise clocks and caches results to skip redundant computation, LVS~\cite{li2018low} pairs feature propagation with an adaptive scheduler, and DVSNet~\cite{xu2018dynamic} synchronizes its segmentation and flow networks. Building on these, TDNet~\cite{Hu2020TemporallyDN}, ETC~\cite{liu2020efficient}, CFFM~\cite{Sun2022CoarsetoFineFM} and MPVSS~\cite{weng2023mask} graft the same tricks onto attention and mask-propagation pipelines through low-resolution feature management and mask reuse. TV3S~\cite{TV3S2025Hesham} takes a different route, processing spatial patches with a Mamba state-space backbone and a hierarchical design that targets accuracy and efficiency together.}

\noindent \revise{\emph{Quantitative trade-off.} \tabref{tab:vss_cityscapes} and \figref{fig:vss_evolution}(a) jointly show that real-time VSS follows two main Pareto routes: lightweight backbones with keyframe pruning maximize speed, while attention-based feature reuse sacrifices some throughput for stronger accuracy. This separates methods optimized for strict latency from those targeting a more balanced accuracy--efficiency trade-off.}

\para{Semi-/Weakly-supervised Methods.} {When labeled video is scarce, this family leans on temporal coherence and spatio-temporal correlation as proxy supervision from unlabeled or sparsely labeled clips~\cite{zhu2019improving,zhuang2022semi,zhuang2024infer}. Naive-Student~\cite{Chen2020NaiveStudentLS} self-trains iteratively, refining pseudo-labels alongside the annotated data, DepthMix~\cite{Hoyer2020ThreeWT} squeezes more from each unlabeled sample through depth mixing and distillation, and SSLTM~\cite{Lao2023SimultaneouslySA} uses attention to model complex spatio-temporal relations while folding in unlabeled data.}

\noindent \revise{\emph{Pseudo-label noise and training instability.} The shared self-training loop carries two systemic risks that video makes worse. First, \textit{temporal amplification of pseudo-label noise}: single-frame errors propagate and get repeatedly self-confirmed across adjacent frames, hitting rare classes and motion boundaries hardest. Naive-Student~\cite{Chen2020NaiveStudentLS} shows this directly, with much smaller gains on long-tailed classes than on head classes, and DepthMix~\cite{Hoyer2020ThreeWT} reports boundary drift once reliable depth priors are missing. Second, \textit{unstable training dynamics}: pseudo-labels drift as the teacher updates, so joint teacher--student optimization is prone to feature and class collapse. Common mitigations (confidence thresholding, EMA teachers, consistency regularization~\cite{zhuang2022semi,zhuang2024infer}) carry their own hyper-parameter sensitivity. Label efficiency in semi-/weakly-supervised VSS therefore comes at a cost, since it shifts statistical risk from annotation to training and leaves a residual mIoU gap to the fully-supervised upper bound, one that narrows as the labeled budget grows but widens with class long-tailedness.}

\para{Common Problem.} \revise{Three coupled issues dominate VSS: temporal consistency, boundary flicker, and the accuracy--latency trade-off. Unlike per-frame segmentation, VSS has to hold pixel labels steady across time, since naive frame-wise inference produces visible flicker at object boundaries and on small or fast-moving regions, the very thing the video-consistency metric mVC$_n$ on VSPW~\cite{miao2021vspw} penalizes. The two dominant remedies pull in opposite directions. Flow warping~\cite{Gadde2017SemanticVC,Jain2018AccelAC} is geometrically interpretable but breaks under large displacement and occlusion, whereas cross-frame attention~\cite{Hu2020TemporallyDN,Sun2022CoarsetoFineFM} is more flexible but scales quadratically with the spatio-temporal volume. This ties temporal stability directly to the accuracy--latency trade-off, since the keyframe scheduling and feature-reuse tricks that restore real-time behavior also weaken cross-frame coherence, a concrete instance of the trilemma in \secref{sec:cross-task}.}

\para{Comparison.} \revise{Three trade-offs structure the VSS design space.
The motion-prior axis pits the geometric explicitness of optical flow~\cite{Gadde2017SemanticVC,Jain2018AccelAC,Ding2019EveryFC} (interpretable and cheap but flow-bounded and fragile under large displacement or illumination shift) against the data-driven flexibility of attention~\cite{Hu2020TemporallyDN,Sun2022CoarsetoFineFM,sun2022mining}, which drops that prior at $\mathcal{O}((THW)^2)$ cost unless mitigated by sparsification or state-space surrogates such as TV3S~\cite{TV3S2025Hesham}.
The fidelity--latency axis is governed less by backbone than by how computation is amortized over time, so keyframe scheduling~\cite{li2018low,xu2018dynamic,weng2023mask,yang2023pruning} and feature distillation~\cite{liu2020efficient,Yang2022CrossImageRK} graft real-time behavior onto heavy backbones.
The supervision axis is orthogonal, semi-/weakly-supervised regimes~\cite{Chen2020NaiveStudentLS,Hoyer2020ThreeWT,Lao2023SimultaneouslySA,zhuang2024infer} attack the annotation bottleneck, not the architecture, paying a residual mIoU gap that widens with class imbalance and pseudo-label noise.}

\begin{table*}[!t]
\centering
\setlength{\tabcolsep}{1.5mm}
\caption{\textbf{Summary of essential characteristics for reviewed VIS methods.}}
\vspace{-10pt}
\label{table:VIS_methods_summary}
\resizebox{0.92\linewidth}{!}{%
\begin{tabular}{c|c|c|c|c}
\hline
\textbf{Methods} & \textbf{Publication} & \textbf{Core Architectures} & \textbf{Technical Features} & \textbf{Training Datasets} \\
\hline
MaskTrack R-CNN \cite{Yang2019VideoIS} & ICCV'2019 & Mask R-CNN & Tracking by detection & YouTube-VIS \cite{Yang2019VideoIS}\\
STEm-Seg \cite{athar2020stem} & ECCV'2020 & FCN & Spatio-temporal embedding & DAVIS \cite{pont20172017}/YouTube-VIS \cite{Yang2019VideoIS}/KITTI-MOTS \cite{voigtlaender2019mots}\\
MaskProp \cite{bertasius2020classifying} & CVPR'2020 & Mask R-CNN & Instance feature propagation & YouTube-VIS \cite{Yang2019VideoIS}\\
MVAE \cite{lin2020video} & CVPR'2020 & Mask R-CNN+VAE & Variational inference & YouTube-VIS \cite{Yang2019VideoIS}/KITTI-MOTS \cite{voigtlaender2019mots}\\
MOTSNet \cite{Porzi2019LearningMT} & CVPR'2020 & Mask R-CNN & Unsupervised learning & KITTI-MOTS \cite{voigtlaender2019mots}/BDD100K \cite{yu2020bdd100k}\\
SemiTrack \cite{Fu2021LearningTT} & CVPR'2021 & SOLO & Semi-supervised learning & Cityscapes \cite{cordts2016cityscapes}/YouTube-VIS \cite{Yang2019VideoIS}\\
CompFeat \cite{fu2021compfeat} & AAAI'2021 & Mask R-CNN & Spatio-temporal feature alignment & YouTube-VIS \cite{Yang2019VideoIS}\\
IFC \cite{hwang2021video} & NIPS'2021 & FCN + Transformer & Inter-frame communication & YouTube-VIS \cite{Yang2019VideoIS}\\
Propose-Reduce \cite{lin2021video} & ICCV'2021 & Mask R-CNN & Propose and reduce & DAVIS \cite{pont20172017}/YouTube-VIS \cite{Yang2019VideoIS}\\
SG-Net \cite{liu2021sg} & CVPR'2021 & FCOS & Single-stage segmentation & YouTube-VIS \cite{Yang2019VideoIS}\\
fIRN \cite{Liu2021WeaklySI} & CVPR'2021 & Mask R-CNN & Weakly-supervised learning & Cityscapes \cite{cordts2016cityscapes}/YouTube-VIS \cite{Yang2019VideoIS}\\
VisTR \cite{wang2021end} & CVPR'2021 & Transformer & Transformer-based segmentation & YouTube-VIS \cite{Yang2019VideoIS}\\
TraDeS \cite{Wu2021TrackTD} & CVPR'2021 & FCN & Tracking by detection & YouTube-VIS \cite{Yang2019VideoIS}/KITTI-MOTS \cite{voigtlaender2019mots}\\
CrossVIS \cite{yang2021crossover} & ICCV'2021 & FCN & Dynamic convolution & YouTube-VIS \cite{Yang2019VideoIS}/OVIS \cite{Qi2021OccludedVI}\\
TPR \cite{li2022improving} & PAMI'2022 & FCN & Temporal pyramid routing & YouTube-VIS \cite{Yang2019VideoIS}\\
VISOLO \cite{han2022visolo} & CVPR'2022 & Transformer & Grid-based spatio-temporal aggregation & COCO \cite{lin2014microsoft}/YouTube-VIS \cite{Yang2019VideoIS}\\
MinVIS \cite{huang2022minvis} & NIPS'2022 & Transformer & Query-based tracking & YouTube-VIS \cite{Yang2019VideoIS}/OVIS \cite{Qi2021OccludedVI}\\
STC \cite{Jiang2022STCSC} & ECCV'2022 & FCN & Contrastive learning & YouTube-VIS \cite{Yang2019VideoIS}/OVIS \cite{Qi2021OccludedVI}\\
TrackFormer \cite{meinhardt2022trackformer} & CVPR'2022 & FCN + Transformer & Track query & MOT17 \cite{milan2016mot16}/MOTS20 \cite{voigtlaender2019mots}\\
SeqFormer \cite{wu2022seqformer} & ECCV'2022 & Transformer & Sequential Transformer & YouTube-VIS \cite{Yang2019VideoIS}\\
IDOL \cite{wu2022defense} & ECCV'2022 & Transformer & Contrastive learning & YouTube-VIS \cite{Yang2019VideoIS}\\
TeViT \cite{yang2022temporally} & CVPR'2022 & Transformer & Pyramid feature interaction & YouTube-VIS \cite{Yang2019VideoIS}/OVIS \cite{Qi2021OccludedVI}\\
GenVIS \cite{heo2023generalized} & CVPR'2023 & Transformer & Query-based training pipeline & YouTube-VIS \cite{Yang2019VideoIS}/OVIS \cite{Qi2021OccludedVI}\\
VideoCutLER \cite{Wang2023VideoCutLERSS} & CVPR'2023 & DINO + FCN & Unsupervised learning & DAVIS \cite{pont20172017}/YouTube-VIS \cite{Yang2019VideoIS}\\
CTVIS \cite{ying2023ctvis} & ICCV'2023 & Transformer & Contrastive learning & YouTube-VIS \cite{Yang2019VideoIS}/OVIS \cite{Qi2021OccludedVI}\\
DVIS \cite{zhang2023dvis} & ICCV'2023 & Transformer & Decoupling strategy & YouTube-VIS \cite{Yang2019VideoIS}/OVIS \cite{Qi2021OccludedVI}\\
OVFormer \cite{fang2024unified} & ECCV'2024 & Transformer & Contrastive learning & LV-VIS \cite{wang2023towards}/YouTube-VIS \cite{Yang2019VideoIS}/OVIS \cite{Qi2021OccludedVI}\\
OV2Seg+ \cite{wang2024ov} & IJCV'2024 & Transformer & Open-vocabulary learning & LVIS \cite{gupta2019lvis}/LV-VIS \cite{wang2023towards}/YouTube-VIS \cite{Yang2019VideoIS}\dots\\
\hline
\end{tabular}}
\vspace{-15pt}
\end{table*}

\subsection{Video Instance Segmentation}
{VIS, introduced in~\cite{Yang2019VideoIS}, unifies per-frame instance detection, segmentation, and cross-frame tracking. We group methods along three frequently overlapping axes: tracking-based methods that explicitly associate detections, attention-based methods that use query or memory mechanisms for set prediction, and semi-/weakly-supervised methods that reduce video-mask annotation cost. The ordering below follows the field's chronology from explicit association to query-based temporal modeling, while noting that individual methods may sit on multiple axes.} \revise{\tabref{table:VIS_methods_summary} summarizes the essential characteristics of the reviewed VIS methods.}

\para{Tracking-based Methods.} {This family lifts ``per-frame detect/segment $+$ temporal association'' from image to video~\cite{bertasius2020classifying,lin2020video,fu2021compfeat,lin2021video,liu2021sg,yang2021crossover,li2022improving}. Chronologically, it moves through three association granularities. \textit{(i) Explicit per-frame}: MaskTrack R-CNN~\cite{Yang2019VideoIS} adds a tracking branch to Mask R-CNN; MOTS~\cite{voigtlaender2019mots} couples pixel masks with multi-object tracking; TraDeS~\cite{Wu2021TrackTD} infers tracking offsets through a cost volume. \textit{(ii) Clip-level implicit}: StemSeg~\cite{athar2020stem} treats the clip as a 3D volume; IFC~\cite{hwang2021video} performs two-stage inter-frame communication before a lightweight class/mask decoder. \textit{(iii) Explicit spatio-temporal coherence}: STC~\cite{Jiang2022STCSC} models spatio-temporal consistency for dynamic motion. The family therefore runs from frame-wise matching to clip-level modeling to long-term consistency, yet ID switches persist under long occlusion and when new instances appear.}

\para{Attention-based Methods.} {In VIS, attention folds detection, segmentation and association into a single set-prediction objective, so methods differ mainly in \emph{how temporal consistency is enforced}, splitting into three groups. \textit{(i) Set-prediction unification}: VisTR~\cite{wang2021end} treats the whole clip as one spatio-temporal volume in which queries attend across frames and a 3D-conv head emits the mask sequence; SeqFormer~\cite{wu2022seqformer} decouples temporal aggregation from per-frame spatial attention to preserve single-frame fidelity; TeViT~\cite{yang2022temporally} builds a multi-scale Transformer pyramid attended by random queries. \textit{(ii) Decoupled task pipelines}: MinVIS~\cite{huang2022minvis} trains at image level and associates instances at inference via cosine similarity of frozen queries; DVIS~\cite{zhang2023dvis} splits the model into detection, segmentation and association streams, cutting cost at the price of externalizing temporal consistency to a lightweight associator. \textit{(iii) Memory and contrastive temporal stability}: VISOLO~\cite{han2022visolo} performs grid-level memory matching; GenVIS~\cite{heo2023generalized} samples diverse multi-clip windows to emulate long-horizon variability; CTVIS~\cite{ying2023ctvis} aligns training and inference via a momentum-updated contrastive query bank. The three groups together trace a unify--decouple--restabilize arc, mirroring the trend in VSS attention-based methods.}

\para{Semi-/Weakly-supervised Methods.} {Mirroring the VSS counterparts, semi-/weakly-supervised VIS exploits temporal coherence as free supervision. SemiTrack~\cite{Fu2021LearningTT} jointly trains instance tracking on images and unlabeled videos; fIRN~\cite{Liu2021WeaklySI} uses motion and temporal consistency as weak supervision; VideoCutLER~\cite{Wang2023VideoCutLERSS} runs a fully unsupervised pipeline (per-frame unsupervised segmentation $\to$ video synthesis $\to$ video-segmenter training); MOTSNet~\cite{Porzi2019LearningMT} auto-generates MOTS data and refines existing methods. Accuracy still trails the supervised ceiling, bottlenecked by pseudo-label drift and occlusion ambiguity, the annotation edge of the trilemma formalized in \secref{sec:cross-task}.}

\para{Common Problem.} \revise{Three intertwined failure modes dominate VIS: long-term tracking, identity switch and occlusion. Together they drive the residual gap on long or occluded clips, most visibly on OVIS~\cite{Qi2021OccludedVI}. \textit{(i) Identity switch:} when two same-class instances cross in the image plane, raw query-similarity association (as in MinVIS) tends to mis-bind the following frames, so frame-level mAP is barely dented while identity-aware metrics fall sharply; CTVIS~\cite{ying2023ctvis} (contrastive query memory) and DVIS~\cite{zhang2023dvis} (explicit association head) are the most direct counter-measures, but each adds hyper-parameters (contrastive temperature, association threshold) whose cross-dataset stability is itself open. \textit{(ii) Long-term tracking:} YouTube-VIS clips are already saturated by query-based models, yet accuracy drops sharply on OVIS, whose videos run an order of magnitude longer than YouTube-VIS clips, since clip-level attention has a temporal receptive field bounded by training clip length. GenVIS~\cite{heo2023generalized} (multi-clip training) and SeqFormer~\cite{wu2022seqformer} attack this from two angles, but neither resolves the underlying need for a \emph{persistent instance state} at inference time. \textit{(iii) Occlusion:} OVIS is curated to expose occlusion, and accuracy on its heavily-occluded subset stays far below the lightly-occluded one. The dominant failure is that an instance vanishes under occlusion and is re-emitted with a new identity, so occlusion collapses into an identity switch. The three failures share one root cause: no persistent, explicit per-instance memory, so temporal consistency must \emph{emerge} inside an attention window, which \secref{sec:cross-task} formalizes as the capacity edge of the trilemma.}

\para{Comparison.} \revise{The architectural axis traces a migration from explicit tracking-by-detection cascades~\cite{Yang2019VideoIS,Wu2021TrackTD} through proposal-free spatio-temporal volumes~\cite{athar2020stem,hwang2021video} to fully unified query Transformers~\cite{wang2021end,huang2022minvis,zhang2023dvis,ying2023ctvis}, trading inductive bias for representational generality. Cascades are modular and cheap but propagate detect--segment--associate errors and falter under occlusion, whereas query Transformers absorb the three sub-tasks into one set-prediction objective at the price of heavy pre-training, quadratic spatio-temporal attention and unstable optimization on long clips. The decisive factor in recent results is no longer raw capacity but \emph{how temporal consistency is enforced}: contrastive memory (CTVIS~\cite{ying2023ctvis}), task decoupling (DVIS~\cite{zhang2023dvis}) or implicit query coherence (MinVIS~\cite{huang2022minvis}). The orthogonal supervision axis targets annotation cost rather than architecture~\cite{Fu2021LearningTT,Liu2021WeaklySI,Wang2023VideoCutLERSS}, and its persistent gap to supervised SOTAs on OVIS~\cite{Qi2021OccludedVI} shows that occlusion robustness, not capacity, is the binding constraint going forward.}

\begin{table*}[!t]
\centering
\setlength{\tabcolsep}{1.5mm}
\caption{\textbf{Summary of essential characteristics for reviewed VPS methods.}}
\vspace{-10pt}
\label{table:VPS_methods_summary}
\resizebox{0.92\linewidth}{!}{%
\begin{tabular}{c|c|c|c|c}
\hline
\textbf{Methods} & \textbf{Publication} & \textbf{Core Architectures} & \textbf{Technical Features} & \textbf{Training Datasets} \\
\hline
VPSNet \cite{Kim2020VideoPS} & CVPR'2020 & Mask R-CNN & Spatio-temporal feature alignment & VIPER-VPS \cite{Kim2020VideoPS}/Cityscapes-VPS \cite{Kim2020VideoPS}\\
ViP-DeepLab \cite{Qiao2020ViPDeepLabLV} & CVPR'2021 & FCN & Depth-aware panoptic segmentation & Cityscapes-VPS \cite{Kim2020VideoPS}\\
SiamTrack \cite{woo2021learning} & CVPR'2021 & Siamese FCN & Supervised contrastive learning & VIPER-VPS \cite{Kim2020VideoPS}/Cityscapes-VPS \cite{Kim2020VideoPS}\\
PVPS \cite{Mei2022WaymoOD} & ECCV'2022 & FCN & Panoramic VPS & WOD: PVPS \cite{Mei2022WaymoOD}\\
Video K-Net \cite{li2022video} & CVPR'2022 & Transformer & Query-based learning & KITTI \cite{geiger2012KITTI}/Cityscapes \cite{cordts2016cityscapes}/VIPSeg \cite{Miao2022LargescaleVP} \\
Clip-PanoFCN \cite{Miao2022LargescaleVP} & CVPR'2022 & FCN & Two-stage segmentation & VIPSeg \cite{Miao2022LargescaleVP}\\
TubeLink \cite{li2023tube} & ICCV'2023 & Transformer & Query-based learning & VIPSeg \cite{Miao2022LargescaleVP} \\
PolyphonicFormer \cite{yuan2022polyphonicformer} & ECCV'2022 & Transformer & Query-based learning & KITTI \cite{geiger2012KITTI}/Cityscapes-VPS \cite{Kim2020VideoPS}\\
ODISE \cite{xu2023open} & CVPR'2023 & UNet + Transformer & Open-vocabulary panoptic features & ADE20K \cite{zhou2019semantic}/COCO \cite{caesar2018coco}\\
\hline
\end{tabular}}
\vspace{-15pt}
\end{table*}

\subsection{Video Panoptic Segmentation}
{VPS~\cite{Kim2020VideoPS} fuses VSS and VIS into one task, where every pixel receives a semantic label while foreground instances are simultaneously distinguished and tracked over time.}

\revise{We organize VPS methods along two orthogonal axes. The \emph{architectural axis} contrasts Dual-branch CNNs with Query-based Transformers, tracing how the field moved from separate stuff/thing streams toward unified set prediction. The \emph{task-setting axis} separates standard 2D VPS from Depth-aware 3D VPS, where depth enters as an input or output signal rather than marking a new architectural family. The two axes do not overlap, since depth-aware VPS runs on either a CNN, as in ViP-DeepLab~\cite{Qiao2020ViPDeepLabLV}, or a query Transformer, as in PolyphonicFormer~\cite{yuan2022polyphonicformer}. We therefore treat depth-aware VPS not as a third paradigm beside Query-based and Dual-branch methods, but as an extension of the task setting. \tabref{table:VPS_methods_summary} summarizes the essential characteristics of the reviewed VPS methods.}

\para{Query-based Methods.} {Query-based VPS recasts segmentation and tracking as a query-interaction problem, removing handcrafted tracking heuristics~\cite{woo2021learning}. Video K-Net~\cite{li2022video} extends K-Net with a set of learnable convolutional kernels that jointly encode semantic categories and instance identities; the kernels interact with pixel features and implicitly maintain temporal consistency. Tube-Link~\cite{li2023tube} chops the video into short clips, generates spatio-temporal tube masks per clip with a Transformer, and links them via inter-query attention, obviating an external tracker. PolyphonicFormer~\cite{yuan2022polyphonicformer} extends the query design with a depth branch, exploiting mutual reinforcement between panoptic segmentation and depth.}

\revise{\emph{Drivers of the query-based shift.} The VPS migration from dual-branch CNNs to query Transformers is driven by three factors. \textit{(i) Unified thing/stuff representation}: dual-branch designs (VPSNet~\cite{Kim2020VideoPS}, PVPS~\cite{Mei2022WaymoOD}) train two streams and stitch them via heuristic fusion, incurring the boundary misalignment and ID drift that query designs were engineered to remove. The query-based turn was consolidated by Video Mask2Former, the video extension of Mask2Former~\cite{cheng2022masked}, which recasts video segmentation as mask classification with a shared Transformer decoder, and by Video K-Net~\cite{li2022video}, which re-represents thing and stuff with a single set of kernels so that fusion becomes an in-pipeline operation. \textit{(ii) Implicitization of association}: the two VIS-validated recipes, image-level training with inference-time query-similarity association (MinVIS~\cite{huang2022minvis}) and explicit decoupling of detection/segmentation/association (DVIS~\cite{zhang2023dvis}), have since been ported to VPS, reducing association from an architectural component to an inference strategy and lowering training cost. \textit{(iii) Temporal stability}: pure query-similarity association breaks under long or occluded clips (cf.\ the VIS long-term-tracking discussion in \secref{sec:method}); CAVIS~\cite{lee2025cavis} carries the CTVIS recipe into VPS via contrastive query memory and an explicit association head, so VPS has now followed the same unify--decouple--restabilize progression previously observed in VIS.}

\para{Dual-branch Methods.} {Dual-branch methods decouple panoptic prediction into a stuff-segmentation branch and a thing-instance branch, fusing the two outputs into a unified panoptic map. PVPS~\cite{Mei2022WaymoOD} extends this design to VPS by stitching five synchronized cameras into a 220$^\circ$ field of view and projecting 2D pixels into a shared 3D space using known camera parameters, achieving consistent segmentation across wide-angle scenes with complex layouts and occlusions.}

\para{Depth-aware Extensions.} {Unlike the query-based and dual-branch paradigms above, this category is \emph{not} an extra architectural paradigm but a 3D extension of the VPS task setting: depth is added to the input/output channels while the underlying architecture can be either CNN- or query-Transformer-based, leaving this axis orthogonal to the paradigm axis. ViP-DeepLab~\cite{Qiao2020ViPDeepLabLV} approaches the task as monocular depth estimation coupled with semantic and instance labeling, recovering 3D point clouds with per-point class and identity. PolyphonicFormer~\cite{yuan2022polyphonicformer} instead adopts a query-based formulation in which a depth head and a panoptic head reinforce each other within a shared decoder.}

\para{Common Problem.} \revise{Three coupled problems dominate VPS: thing/stuff balance, long-tail coverage, and the real-time bottleneck. What sets VPS apart from VSS and VIS is not that it stacks the two tasks, but that it must reconcile thing instances and stuff regions inside one non-overlapping video-level output. The benchmark trend in \tabref{tab:vps} points to three persistent bottlenecks. \textit{(i) Thing/stuff imbalance:} query-based methods share representation capacity across instance-level things and dense stuff regions, so balancing identity-aware instance reasoning against stable background parsing stays hard. Dual-branch designs ease this competition but bring boundary-fusion errors of their own. \textit{(ii) Long-tail category coverage:} accuracy climbs more steadily on compact urban benchmarks than on broader in-the-wild settings such as VIPSeg~\cite{Miao2022LargescaleVP}, which shows that vocabulary expansion and data imbalance increasingly cap progress. \textit{(iii) Real-time VPS:} latency stays under-reported and immature as an evaluation axis, since online VPS must hold thing identities while keeping dense panoptic predictions. This sharpens the accuracy--latency trade-off beyond what VSS faces and ties VPS straight to the cross-task bottlenecks in \secref{sec:cross-task}.}

\para{Comparison.} \revise{The field's pivot from dual-branch CNNs~\cite{Kim2020VideoPS,Mei2022WaymoOD} toward query-based Transformers~\cite{li2022video,li2023tube} is driven less by backbone capacity than by the representational economy of \emph{set prediction}: a single set of learnable kernels or queries can carry both semantic class and instance identity, dissolving the brittle thing/stuff fusion heuristics of two-branch designs and rendering cross-frame association a by-product of query persistence.
Depth-awareness is more accurately understood as a \emph{task extension} (a 3D-grounded variant of VPS instantiated on either CNN backbones~\cite{Qiao2020ViPDeepLabLV} or query Transformers~\cite{yuan2022polyphonicformer}) rather than as a parallel methodological category.
These advances are not without cost: query attention scales with $\mathcal{O}(N_q\!\cdot\!T\!\cdot\!HW)$, real-time VPS remains an open problem, and the contrast between near-saturated results on Cityscapes-VPS and the much lower scores on VIPSeg~\cite{Miao2022LargescaleVP} (124 categories, long-tailed) shows that the binding constraint has shifted from architectural expressiveness to long-tail coverage and stable thing/stuff balance under data scarcity.}

\begin{table*}[!t]
\centering
\setlength{\tabcolsep}{1.5mm}
\caption{\textbf{Summary of essential characteristics for reviewed VTS methods.}}
\vspace{-10pt}
\label{table:VTS_methods_summary}
\resizebox{0.92\linewidth}{!}{%
\begin{tabular}{c|c|c|c|c}
\hline
\textbf{Methods} & \textbf{Publication} & \textbf{Core Architectures} & \textbf{Technical Features} & \textbf{Training Datasets} \\
\hline
Track R-CNN \cite{voigtlaender2019mots} & CVPR'2019 & Mask R-CNN & Tracking by detection & KITTI MOTS \cite{voigtlaender2019mots}/MOTSChallenge \cite{voigtlaender2019mots}\\
PointTrack \cite{xu2020segment} & ECCV'2020 & PointTrack & Tracking by points & APOLLO MOTS \cite{xu2020segment} \\
MaskProp \cite{bertasius2020classifying}  & CVPR'2020 & Mask R-CNN & Instance feature propagation & YouTube-VIS \cite{Yang2019VideoIS}\\
ASB \cite{choudhuri2021assignment} & ICCV'2021 & Assignment Space & Top-k + Dynamic programming & KITTI MOTS \cite{voigtlaender2019mots}/MOTSChallenge \cite{voigtlaender2019mots}\\
MPNTrackSeg \cite{braso2022multi} & IJCV'2022 & MPN + GNN & Graph-based & MOT20 \cite{dendorfer2020mot20}/KITTI \cite{geiger2012KITTI}/HiEve \cite{lin2020human}\\
MITS \cite{xu2023integrating} & ICCV'2023 & Transformer & Joint tracking and segmentation & YouTube-VOS \cite{xu2018youtube}/DAVIS \cite{perazzi2016benchmark}\\
SAMTrack \cite{cheng2023segment} & arXiv'2023 & SAM + DeAOT & Multimodal interaction & DAVIS \cite{perazzi2016benchmark,pont20172017}\\
SAM-PT \cite{rajivc2023segment} & arXiv'2023 & SAM + PT & Tracking by points & DAVIS \cite{perazzi2016benchmark}/YouTube-VOS \cite{xu2018youtube}/BDD100K \cite{yu2020bdd100k}\\
DEVA \cite{cheng2023tracking} & ICCV'2023 & SAM & Decoupled segmentation and tracking & VIPSeg \cite{Miao2022LargescaleVP}/YouTube-VIS \cite{Yang2019VideoIS} \\
SAM 2 \cite{ravi2024sam} & arXiv'2024 & SAM  & Prompt segmentation & SA-V \cite{ravi2024sam} \\
\hline
\end{tabular}}
\vspace{-15pt}
\end{table*}

\subsection{Video Tracking \& Segmentation}
{VTS jointly detects, tracks and pixel-segments objects over time, with applications in autonomous driving, surveillance and action recognition. We organize methods by how an instance is represented across time (Point-based, Two-stage proposal-based and Efficient assignment / decoupled), and the Comparison below revisits them along two orthogonal axes, \emph{coupling} (tightly-coupled vs.\ decoupled) and \emph{representation} (dense vs.\ sparse).} \revise{\tabref{table:VTS_methods_summary} summarizes the essential characteristics of the reviewed VTS methods.}

\para{Point-based Methods.} {Point-based methods replace dense convolutional features with sparse 2D point clouds to better separate foreground from background. PointTrack~\cite{xu2020segment} represents both foreground and surrounding regions as point clouds across four modalities (offset, color, category, position) for joint segmentation and tracking. SAM-PT~\cite{rajivc2023segment} instead propagates positive and negative points across frames and refines masks iteratively through SAM~\cite{kirillov2023segment}, discarding occluded points and adding newly visible ones. MITS~\cite{xu2023integrating} pushes localization further, coupling segmentation masks with bounding boxes through self-attention for unified tracking. Across the three, point-level representations reach competitive accuracy at a lightweight cost.}

\para{Two-stage Methods.} {Two-stage methods generate region proposals first and only then classify, segment and associate across frames, so localization stays decoupled from refinement. Track R-CNN~\cite{voigtlaender2019mots} sets the pattern, augmenting a ResNet-101 backbone with 3D convolutions and an association head that matches embeddings across frames. MaskProp~\cite{bertasius2020classifying} extends Mask R-CNN along the same route, adding a mask-propagation branch that carries instance-specific features through time. MPNTrackSeg~\cite{braso2022multi} reframes the association step as graph message passing under flow-conservation constraints, classifying association edges and predicting masks from CNN embeddings. SAM-Track~\cite{cheng2023segment} widens the setting, supporting both interactive (SAM~\cite{kirillov2023segment} $+$ Grounding DINO~\cite{liu2024grounding}) and autonomous modes to curb identity drift in unconstrained scenes.}

\para{Efficient Methods.} {Efficient methods cut the cost of association by recasting it as an optimization or by splitting the pipeline apart. ASB~\cite{choudhuri2021assignment} takes the optimization route, reformulating tracking as assignment-space search, where top-$k$ frame-to-frame matches are found via Hungarian--Murty under a cost function that blends IoU, appearance and distance, with dynamic programming handling global optimization and long-distance re-association across interrupted detections. DEVA~\cite{cheng2023tracking} takes a decoupling route, separating segmentation from tracking so task-specific segmenters run per frame while bidirectional propagation module links them over time and generalizes across tasks.}

\para{Common Problem.} \revise{In VTS the binding constraint is duration, since holding an instance together across a long stream is harder than parsing any single clip. VTS and VIS share the same output structure $\{(M_i, c_i, \mathrm{id}_i)\}$ (\secref{sec:def}), yet they stress different temporal regimes. VIS is usually scored on short clips, so it leans on clip-level set prediction and query-based attention, whereas VTS targets longer streams where identity continuity, occlusion recovery and frame-to-frame association dominate. The metric emphasis moves with them, since VIS mainly reports AP while VTS foregrounds tracking-aware measures such as sMOTSA, MOTSA and IDS. Because of this, VTS methods tend to favor lightweight representations, explicit association, graph matching or propagation-based designs rather than fully porting VIS architectures. So the divergence between VIS and VTS is driven less by output format than by temporal horizon, evaluation protocol and deployment scenario.}

\para{Comparison.} \revise{The defining VTS question is not \emph{what} to segment but \emph{how} segmentation and tracking are coupled. Along the coupling axis, tightly-coupled joint frameworks~\cite{voigtlaender2019mots,xu2023integrating} learn embeddings and association end-to-end but turn brittle once prolonged occlusion perturbs the embedding manifold. Decoupled pipelines~\cite{cheng2023tracking,rajivc2023segment,cheng2023segment} adopt the opposite strategy, handing per-frame segmentation to SAM~\cite{kirillov2023segment} and leaving temporal linking to a lightweight propagator, so they inherit open-vocabulary generality yet still struggle with long-range re-ID. Along the representation axis, point-based encodings~\cite{xu2020segment,rajivc2023segment} trade dense pixel coverage for a sparse, deformation-robust proxy, whereas assignment-space~\cite{choudhuri2021assignment} and graph message-passing formulations~\cite{braso2022multi} recast tracking as combinatorial optimization at extra inference cost. PointTrack~\cite{xu2020segment} still remains competitive on KITTI-MOTS (\tabref{tab:vts}) despite a six-year gap, which suggests that in VTS the principal driver of accuracy is careful representational design rather than raw capacity or foundation-model scale.}

\subsection{Open-Vocabulary Video Segmentation}
{OVVS fuses semantic segmentation, temporal tracking, and language-based recognition, then leans on vision--language models to reach past a fixed vocabulary. Three challenges remain unresolved: language ambiguity, temporal alignment drift, and rare-class generalization.}

{OVVS grafts open-vocabulary recognition onto a host VSP task. \textit{(VIS)} OVFormer~\cite{fang2024unified} aligns vision--language features with instance queries via unified embedding alignment and semi-online inference; OV2Seg+~\cite{wang2024ov} pairs a universal proposal generator with a memory-induced tracker and a VLM-driven classifier; CLIP-VIS~\cite{zhu2025clip} adapts a frozen CLIP backbone through class-agnostic mask generation and temporal top-$K$ query matching. \textit{(VPS)} ODISE~\cite{xu2023open} performs image-level open-vocabulary panoptic segmentation from text-to-image diffusion features. We list it as a component that can supply open-vocabulary panoptic recognition to a video pipeline, not as a native temporal VPS model. \textit{(VSS)} OV2VSS~\cite{li2024towards} adds a spatio-temporal fusion module, random-frame augmentation and a dedicated video--text encoder. All three share one blueprint: delegate recognition to a frozen vision--language backbone, and leave only a lightweight temporal or alignment adapter trainable.}

\para{Common Problem.} \revise{The binding constraint in OVVS is recognition, not mask generation. Since every current pipeline delegates classification to a frozen image-level vision--language backbone~\cite{radford2021learning} and trains only lightweight adapters, three failure modes recur. \textit{(i) Granularity ambiguity:} the same segment can legitimately match several vocabulary levels (``vehicle''/``car''/``sedan''), so reported scores partly depend on how the test vocabulary is phrased (\secref{sec:def}). \textit{(ii) Semantic flicker:} because the vision--language embedding is trained on still images, the same object can drift across textual neighborhoods under pose or illumination change, violating the class-consistency constraint that closed-set VSP takes for granted. \textit{(iii) Coupled long-tail and zero-shot generalization:} the large gaps to supervised counterparts on VSPW (\tabref{tab:vss_vspw}) and YouTube-VIS (\tabref{tab:vis_youtube}) persist even with comparable backbones, indicating that open-vocabulary recognition, rather than temporal modeling, is the bottleneck (cf.\ \secref{sec:cross-task}).}

\para{Comparison.} \revise{Current OVVS designs differ mainly in where they place the vision--language boundary. Adapter-style methods (CLIP-VIS~\cite{zhu2025clip}, OV2VSS~\cite{li2024towards}) keep the VLM frozen, preserving zero-shot generality but capping how tightly language can shape the masks; alignment-style methods (OVFormer~\cite{fang2024unified}) train the embedding alignment end-to-end, buying base-class accuracy at the risk of drifting toward seen categories; decoupled pipelines (OV2Seg+~\cite{wang2024ov}, ODISE~\cite{xu2023open}) separate class-agnostic mask generation from VLM classification. Notably, the three routes land within a few AP of one another on YouTube-VIS (\tabref{tab:vis_youtube}) while all trailing supervised counterparts by 20$+$ AP, which suggests that the binding constraint is the frozen image-level VLM itself rather than any particular adapter design.}

\subsection{Cross-task Discussion}\label{sec:cross-task}
\revise{Although VSS, VIS, VPS, VTS and OVVS are usually treated as distinct tasks with their own metrics, datasets and architectures, each must answer the same three questions: \emph{what} a pixel denotes, \emph{which} object it belongs to, and \emph{how} that assignment holds across time. The tasks differ mainly in which question they push to the front. This shared substrate breeds cross-cutting challenges that no single task can settle alone.}

\subsubsection{Per-task Challenge--Solution--Paradigm Map.}
\revise{Before the shared themes, we read each task through one common challenge--solution--paradigm structure. \textbf{VSS} focuses on per-pixel temporal stability (mVC$_n$), moving from flow warping~\cite{Gadde2017SemanticVC} to attention~\cite{Sun2022CoarsetoFineFM} and then to state-space modeling~\cite{TV3S2025Hesham}. \textbf{VIS} focuses on cross-frame identity, progressing from tracking-by-detection~\cite{Yang2019VideoIS} to set prediction~\cite{huang2022minvis,zhang2023dvis} and contrastive query memory~\cite{ying2023ctvis,lee2025cavis}. \textbf{VPS} targets a balanced thing/stuff output with no overlap, spanning dual-branch fusion~\cite{Kim2020VideoPS,Mei2022WaymoOD}, query unification~\cite{li2022video,li2023tube}, and depth-aware 3D extensions~\cite{Qiao2020ViPDeepLabLV,yuan2022polyphonicformer}. \textbf{VTS} focuses on long-horizon identity through joint detection and tracking~\cite{voigtlaender2019mots,xu2023integrating}, decoupled SAM-style propagation~\cite{cheng2023tracking,rajivc2023segment}, and point-cloud or assignment-space embeddings~\cite{xu2020segment,choudhuri2021assignment}. \textbf{OVVS} targets open-set generalization through CLIP alignment~\cite{radford2021learning}, vision--language adapters~\cite{wang2023towards,fang2024unified}, and mask-free text-driven segmentation~\cite{xu2023open}. Across tasks, the recurring pattern is that explicit priors give way to implicit set or query prediction, then to explicit memory or contrastive stabilization.}

\subsubsection{Temporal Consistency.}
\revise{Temporal consistency is the one challenge every VSP task shares, though it differs in each: VSS wants stable per-pixel labels under motion (mVC$_n$ on VSPW~\cite{miao2021vspw}), VIS and VPS want stable identity, and VTS wants long-horizon stability. Methods lean on three mechanisms: explicit flow warping~\cite{Gadde2017SemanticVC,Jain2018AccelAC}, implicit cross-frame attention or memory~\cite{Hu2020TemporallyDN,ying2023ctvis}, and built-in coherence of pre-trained queries~\cite{huang2022minvis,zhang2023dvis}. None universally dominates since flow excels at short range, attention at mid range, and query coherence at long range under a strong foundation. Recent SOTAs therefore \emph{combine} all three, so consistency is best read as a multi-scale property.}

\subsubsection{Identity Preservation under Occlusion.}
\revise{Whenever an object vanishes and returns, every identity-bearing task (VIS, VPS, VTS, instance-level OVVS) must re-identify it against a background that has since moved on. This is the most visible failure in practice, since KITTI-MOTS~\cite{voigtlaender2019mots} ID switches, the YouTube-VIS--OVIS gap~\cite{Qi2021OccludedVI} and steep VPQ decay on long VIPSeg~\cite{Miao2022LargescaleVP} sequences all trace back to one weakness. Two routes answer it: contrastive memory-based association, which trades footprint for re-ID robustness~\cite{ying2023ctvis,wu2022defense}, and SAM-based propagation~\cite{kirillov2023segment}, which delegates identity to a task-agnostic linker~\cite{cheng2023tracking,rajivc2023segment}. Neither fully resolves long-range occlusion, and the short- versus long-clip benchmark gap stays the most reliable diagnostic.}

\subsubsection{Long-tail and Open-world Generalization.}
\revise{As datasets grow from a few urban classes (Cityscapes~\cite{cordts2016cityscapes}, KITTI~\cite{geiger2012KITTI}) to large-vocabulary corpora (VSPW~\cite{miao2021vspw}, VIPSeg~\cite{Miao2022LargescaleVP}, LV-VIS~\cite{wang2023towards}), a long tail appears across all five tasks: rare classes are systematically under-segmented, and unseen ones lie out of reach for closed-vocabulary models. OVVS makes the open-world assumption explicit, yet its mIoU and AP (\tabref{tab:vss_vspw}, \tabref{tab:vis_youtube}) still trail supervised counterparts by a wide margin, which says open-vocabulary recognition is currently \emph{generalization-limited rather than vocabulary-limited}. The shared bottleneck is alignment between mask-trained visual representations and the broader semantic prior held in vision--language foundations~\cite{radford2021learning}, so a gain on any one task (\eg, OVFormer~\cite{fang2024unified}, ODISE~\cite{xu2023open}) should transfer, which argues for a single cross-task agenda.}

\subsubsection{The Annotation--Capacity--Latency Trilemma.}
\revise{Every VSP task is constrained by a three-way tension between annotation cost, capacity and latency. Pixel-accurate, temporally consistent labels cost an order of magnitude more than image masks, hence label-efficient learning recurs throughout VSP: VSS via self-training/consistency~\cite{Chen2020NaiveStudentLS,Hoyer2020ThreeWT,Lao2023SimultaneouslySA}, VIS via pseudo-label propagation/unsupervised synthesis~\cite{Fu2021LearningTT,Liu2021WeaklySI,Wang2023VideoCutLERSS}, and VTS and OVVS via foundation models~\cite{kirillov2023segment,radford2021learning} as zero-/few-shot supervision. Yet capacity gains (wider backbones, longer pre-training, heavier query attention) come at a latency cost (cf.\ \tabref{tab:vss_cityscapes}) incompatible with autonomous driving and embodied perception. The triangle resists edge-by-edge optimization, since data-efficient learning needs capacity to absorb noisy supervision, and real-time deployment needs that capacity distilled without losing temporal coherence. A unified treatment of the trilemma is the most under-explored opening in VSP today.}

\subsubsection{Representative Failure Modes.}\label{sec:failure}
\revise{A handful of concrete failures recur, and each points to a deeper limitation. \textit{(i) Prolonged occlusion:} query-based VIS Transformers (MinVIS~\cite{huang2022minvis}, DVIS~\cite{zhang2023dvis}) drop $>$10 AP on OVIS~\cite{Qi2021OccludedVI} against YouTube-VIS, and SAM-based propagation~\cite{cheng2023tracking,rajivc2023segment} loses identity once the disocclusion gap runs past a few seconds, which shows identity is encoded as a \emph{short-horizon} property. \textit{(ii) Fast motion and motion blur:} flow-based VSS~\cite{Gadde2017SemanticVC,Jain2018AccelAC,Ding2019EveryFC} and tracking-by-detection VIS~\cite{Yang2019VideoIS,Wu2021TrackTD} rest on inter-frame correspondence that collapses under large displacement and blur, producing ghosting and broken tracklets, so attention aggregation eases the symptom, not the cause. \textit{(iii) Identity switch in crowded scenes:} on VIPSeg~\cite{Miao2022LargescaleVP} and BDD100K~\cite{yu2020bdd100k}, decoupled trackers fall back on insufficiently discriminative appearance embeddings, plateauing thing-VPQ below stuff-VPQ. Contrastive embeddings~\cite{ying2023ctvis,wu2022defense} raise the ceiling, though residual error correlates with crowd density. \textit{(iv) Rare and fine-grained categories:} VSPW~\cite{miao2021vspw} (124 classes) and LV-VIS~\cite{wang2023towards} (1{,}196 classes, 553 of which are held out as novel) both exhibit heavily long-tailed category distributions, on which rare classes are systematically under-segmented, since closed-vocabulary training never internalizes the \emph{semantic prior} for rare classes, a gap that current OVVS methods~\cite{wang2023towards,fang2024unified,xu2023open} only partly close.}

\subsubsection{Quantitative Cross-task Synthesis.}
\revise{Reading \tabref{tab:vss_cityscapes}--\tabref{tab:vts} side by side surfaces three patterns. \emph{Architectural convergence:} the accuracy leader of every task except VTS is a Transformer decoder (VPSeg/TubeFormer 82.5/63.2 mIoU, CTVIS 55.1 AP, PolyphonicFormer/TubeLink 65.4 VPQ/49.4 STQ), all of which appeared in 2022--2024, with non-uniform dividends ($\Delta\!+\!25.7/+\!7.4$ VSPW/Cityscapes mIoU, $+\!24.8$ VIS AP, $+\!8.0/+\!17.9$ Cityscapes-VPS/VIPSeg); VTS is the exception, where a point-cloud embedding still delivers $+\!9.3$ sMOTSA and a ${\sim}4\!\times$ IDS reduction over its tracking-by-detection baseline. \emph{Efficiency cliff:} among these per-task accuracy leaders, only PointTrack runs in real time (22.2 FPS), while the VSS/VIS/VPS leaders report single-digit FPS or none; speed-oriented designs such as BLO (30.8 FPS) sit several mIoU below the accuracy frontier, so the trilemma's latency edge binds across tasks. \emph{Annotation ceiling:} open-vocabulary entries trail supervised ones by 46 mIoU on VSPW and 20--23 AP on YouTube-VIS, a 20--46-point gap localizing the bottleneck to labeled supervision, not capacity.}

\revise{The boundaries between VSS, VIS, VPS, VTS and OVVS are therefore largely a benchmark-design artefact. The underlying questions are shared, and progress increasingly comes from methods that start out from this common ground.}

\begin{table*}[!t]
\centering
\setlength{\tabcolsep}{1.2mm}
\caption{{\textbf{Statistics of Video Scene Parsing datasets.}}}
\vspace{-10pt}
\label{tab:dataset}
\resizebox{0.95\linewidth}{!}{%
\label{tab:video_segmentation_datasets}
\renewcommand{\arraystretch}{1.2}
\begin{tabular}{lccccccc}
    \toprule
    \textbf{Datasets} & \textbf{Year} & \textbf{Pub.} & \textbf{\#Videos} & \textbf{Train/Val/Test/Dev} & \textbf{Annotations} & \textbf{Purposes} & \textbf{\#Classes} \\
    \midrule
    CamVid~\cite{brostow2008segmentation} & 2009 & PRL & 4 & (frame: 467/100/233/-) & VSS & Urban & 11 \\
    NYUDv2~\cite{silberman2012indoor} & 2012 & ECCV & 518 & (frame: 795/654/-/-) & VSS & Indoor & 40 \\
    Cityscapes~\cite{cordts2016cityscapes} & 2016 & CVPR & 5,000 & 2,975/500/1,525 & VSS & Urban & 19 \\
    ACDC~\cite{sakaridis2021acdc} & 2021 & ICCV & - & (frame: 1600/406/2000) & VSS & Urban & 19 \\
    VSPW~\cite{miao2021vspw} & 2021 & CVPR & 3,536 & 2,806/343/387/- & VSS & Generic & 124 \\
    MVSeg~\cite{Ji2023MultispectralVS} & 2023 & CVPR & 738 & 452/84/202 & VSS & Urban & 26 \\
    YouTube-VIS~\cite{Yang2019VideoIS} & 2019 & ICCV & 3,859 & 2,985/421/453/- & VIS & Generic & 40 \\
    KITTI MOTS~\cite{voigtlaender2019mots} & 2019 & CVPR & 21 & 12/9/-/- & VIS & Urban & 2 \\
    MOTSChallenge~\cite{voigtlaender2019mots} & 2019 & CVPR & 4 & -/-/-/- & VIS & Urban & 1 \\
    OVIS~\cite{Qi2021OccludedVI} & 2021 & IJCV & 901 & 607/140/154/- & VIS & Generic & 25 \\
    BDD100K~\cite{yu2020bdd100k} & 2020 & ECCV & 100,000 & 7,000/1,000/2,000/- & VSS, VIS & Driving & 40 (VSS), 8 (VIS) \\
    Cityscapes-VPS~\cite{Kim2020VideoPS} & 2020 & CVPR & 500 & 400/100/-/- & VPS & Urban & 19 \\
    VIPER-VPS~\cite{Kim2020VideoPS} & 2020 & CVPR & 124 & (frame: 134K/50K/70K/-) & VPS & Urban & 23 \\
    KITTI-STEP~\cite{Weber2021STEPSA} & 2021 & NeurIPS & 50 & 12/9/29 & VPS & Urban & 19 \\
    MOTChallenge-STEP~\cite{Weber2021STEPSA} & 2021 & NeurIPS & 4 & 2/2/-/- & VPS & Urban & 7 \\
    VIPSeg~\cite{Miao2022LargescaleVP} & 2022 & CVPR & 3536 & 2806/343/387 & VPS & Generic & 124 \\
    WOD: PVPS~\cite{Mei2022WaymoOD} & 2022 & ECCV & 2860 & 2800/20/40 & VPS & Urban & 28 \\
    APOLLO MOTS~\cite{xu2020segment} & 2020 & ECCV & - & (frame: 3.4k/2.3k/5.7k/-) & VIS & Urban & - \\
    HiEve~\cite{lin2020human} & 2023 & IJCV & 32 & 19/0/13/- & VTS & Human & 14 \\
    DAVIS16~\cite{perazzi2016benchmark} & 2016 & CVPR & 50 & 30/20/-/- & VOS (object-level) & Generic & - \\
    DAVIS17~\cite{pont20172017} & 2017 & - & 150 & 60/30/30/30 & VOS (instance-level) & Generic & - \\
    \bottomrule
\end{tabular}}
\vspace{-10pt}
\end{table*}

\section{Datasets and Metrics}
We overview common VSP datasets (\secref{sec:datasets}) and evaluation metrics (\secref{sec:metrics}).
\subsection{Datasets}
\label{sec:datasets}
We organize representative VSP datasets by sub-task with a one-line summary each, and give the full numerics in \tabref{tab:dataset}.

\subsubsection{VSS Datasets}
\begin{itemize}[left=0pt,itemsep=0.5pt,topsep=1pt]
    \item{\textbf{CamVid~\cite{brostow2008segmentation}}: 4 driving videos / 800 labeled frames (467/100/233); the first video dataset with semantic class labels.}
    \item{\textbf{NYUDv2~\cite{silberman2012indoor}}: 1{,}449 densely annotated RGB-D indoor frames from Kinect; class and instance IDs for indoor segmentation.}
    \item{\textbf{Cityscapes~\cite{cordts2016cityscapes}}: stereo driving sequences from 50 cities / 5K fine + 20K coarse pixel-level annotations / 30 classes; the canonical urban VSS benchmark.}
    \item{\textbf{ACDC~\cite{sakaridis2021acdc}}: 4{,}006 adverse-weather images (fog/night/rain/snow) paired with normal-condition counterparts; targets robustness evaluation.}
    \item{\textbf{VSPW~\cite{miao2021vspw}}: 3{,}536 videos / 251{,}632 frames / 124 classes / 15 FPS / 200+ scenes; the first large-scale, multi-scene, densely annotated VSS benchmark.}
    \item{\textbf{MVSeg~\cite{Ji2023MultispectralVS}}: 738 paired RGB-thermal videos / 3{,}545 day-night urban frames / 26 classes; targets multi-spectral VSS.}
\end{itemize}

\subsubsection{VIS Datasets}
\begin{itemize}[left=0pt,itemsep=0.5pt,topsep=1pt]
    \item{\textbf{YouTube-VIS~\cite{Yang2019VideoIS}}: 3{,}859 videos (2{,}985/421/453 splits) / 40 categories / 131K masks; extended from YouTube-VOS, the canonical VIS benchmark.}
    \item{\textbf{KITTI MOTS~\cite{voigtlaender2019mots}}: 21 driving videos / 8{,}008 frames / 26{,}899 cars + 11{,}420 pedestrians; for joint detection-tracking-segmentation.}
    \item{\textbf{MOTSChallenge~\cite{voigtlaender2019mots}}: 4 videos / 2{,}862 frames / 26{,}894 pedestrian masks; a pedestrian-only crowded-scene tracking benchmark.}
    \item{\textbf{OVIS~\cite{Qi2021OccludedVI}}: 901 videos ($\bar{t}\!\approx\!12.8$s) / 25 categories / 296K masks / 5{,}223 instances; targets heavy occlusion.}
\end{itemize}

\subsubsection{VPS Datasets}
\begin{itemize}[left=0pt,itemsep=0.5pt,topsep=1pt]
    \item{\textbf{Cityscapes-VPS~\cite{Kim2020VideoPS}}: 2{,}400/300/300 frames / 6 annotated frames per 30-frame clip / 19 classes; an urban VPS benchmark.}
    \item{\textbf{VIPER-VPS~\cite{Kim2020VideoPS}}: 254K synthetic GTA-V driving frames at $1080{\times}1920$ / 10 thing + 13 stuff classes; a large-scale synthetic VPS source.}
    \item{\textbf{KITTI-STEP~\cite{Weber2021STEPSA}}: 50 videos / 18{,}181 frames / 2 thing + 17 stuff classes / 126K masks; a STEP-protocol urban benchmark.}
    \item{\textbf{MOTChallenge-STEP~\cite{Weber2021STEPSA}}: 4 videos / 2{,}075 frames / 1 thing + 6 stuff classes; a pedestrian STEP benchmark.}
    \item{\textbf{VIPSeg~\cite{Miao2022LargescaleVP}}: 3{,}536 videos / 84{,}750 frames / 232 scenes / 58 thing + 66 stuff classes / 926K masks; the largest in-the-wild VPS benchmark.}
    \item{\textbf{WOD:PVPS~\cite{Mei2022WaymoOD}}: 2{,}860 videos / 100K frames / 8 tracking + 28 semantic classes; multi-camera, temporally consistent panoptic annotations.}
\end{itemize}

\subsubsection{VTS Datasets}
\begin{itemize}[left=0pt,itemsep=0.5pt,topsep=1pt]
    \item{\textbf{APOLLO MOTS~\cite{xu2020segment}}: 11.4K ApolloScape frames / 5.65 cars per frame ($2\times$ KITTI MOTS density, $2.5\times$ crowded cars); car-centric, for 2D/3D tracking.}
    \item{\textbf{HiEve~\cite{lin2020human}}: $1$M pose annotations / 56K complex-event action instances / trajectories averaging 480+ frames; human-centric.}
    \item{\textbf{DAVIS16~\cite{perazzi2016benchmark}}: 50 sequences (30/20) / 3{,}455 frames / single object per clip; the canonical short-clip VOS-style benchmark.}
    \item{\textbf{DAVIS17~\cite{pont20172017}}: 150 sequences / 10{,}459 frames / 376 objects; multi-object clips with distractors, occlusion and fast motion.}
\end{itemize}

\subsection{Metrics}\label{sec:metrics}
{We review the thirteen metrics most commonly used in VSP, ordered from raw overlap through temporal, efficiency and identity-aware scores, and close with per-task reporting recommendation.}

\para{Intersection over Union (IoU).}
IoU is the base template the rest of the section builds on, the Jaccard overlap between a predicted mask and its ground truth:
\begin{equation}
    \text{IoU} = \frac{\text{TP}}{\text{TP} + \text{FP} + \text{FN}},
\end{equation}
with TP/FP/FN the true/false positives and false negatives. IoU jointly penalizes over- and under-segmentation.

\para{Mean Intersection over Union (mIoU).}
Averaging IoU over $C$ classes stops frequent classes from swamping rare ones:
\begin{equation}
    \text{mIoU} = \frac{1}{C} \sum_{i=1}^{C} \text{IoU}_i.
\end{equation}
This class balance is why mIoU is the headline number for VSS.

\para{Video Consistency (VC).}
VC checks whether labels hold steady across $n$ consecutive frames. With $S_i$/$S'_i$ the ground-truth/predicted labels of frame $i$ and $C$ the clip length,
\begin{equation}
    {\rm VC}_n = \frac{1}{C - n + 1}\sum_{i = 1}^{C - n + 1}\frac{\bigl\lvert(\bigcap_{j = 0}^{n - 1}S_{i + j})\cap(\bigcap_{j = 0}^{n - 1}S'_{i + j})\bigr\rvert}{\bigl\lvert\bigcap_{j = 0}^{n - 1}S_{i + j}\bigr\rvert},
\end{equation}
and the dataset-level mean $\text{mVC}_n = \tfrac{1}{N}\sum_{k=1}^{N}\text{VC}_n^{(k)}$. A higher $\text{mVC}_n$ means fewer temporal flickers.

\para{Frames per Second (FPS).} {FPS reports raw throughput, $\text{FPS} = 1/\bar{t}$ with $\bar{t}=\tfrac{1}{N}\sum_{i=1}^{N} t_i$. Being an average, it hides tail latency.}

\para{Max Latency.} {Max Latency catches what FPS hides, the worst single frame $\max_{i}\{t_i\}$. It is the number that matters for hard real-time settings such as autonomous driving or AR, where one spike breaks the perception--control loop.}

\para{Average Precision (AP).}
AP is to VIS what mIoU is to VSS, the image metric lifted to video and averaged over $|T|$ IoU thresholds:
\begin{equation}
    \mathrm{AP} = \frac{1}{\lvert T\rvert}\sum_{t \in T} \mathrm{AP}_t.
\end{equation}
Sweeping thresholds rewards precise masks and confident detections together, which makes AP the headline VIS score on YouTube-VIS and OVIS.

\para{Average Recall (AR).}
The best recall reachable when model returns its top-$K$ instances per video:
\begin{equation}
    \mathrm{AR} = \frac{1}{\lvert V\rvert}\sum_{v \in V} \frac{\text{TP}_v(K)}{\text{GT}_v}.
\end{equation}
AR$_{10}$ pairs with AP as a recall ceiling.

\para{Segmentation and Tracking Quality (STQ).}
STQ splits video tracking into two questions, whether an instance keeps its identity (Association Quality, AQ) and whether its pixels are labeled right (class-level Segmentation Quality, SQ):
\begin{equation}
    \text{AQ} = \frac{1}{\lvert G\rvert} \sum_{z_g\in G} \frac{1}{\lvert gid(z_g)\rvert} \sum_{z_f:\, z_f\cap z_g\neq\emptyset} {\rm TPA}(z_f,z_g)\times {\rm IoU}_{id}(z_f,z_g),
\end{equation}
\begin{equation}
    \text{SQ} = \frac{1}{\lvert C\rvert}\sum_{c\in C}\frac{\lvert f_{\mathrm{sem}}(c)\cap g_{\mathrm{sem}}(c)\rvert}{\lvert f_{\mathrm{sem}}(c)\cup g_{\mathrm{sem}}(c)\rvert},\qquad \text{STQ} = \sqrt{\rm AQ \times SQ},
\end{equation}
the geometric mean fuses both into one score, and AQ read alone surfaces identity errors that frame-level metrics miss.

\para{Multi-Object Tracking and Segmentation Precision (MOTSP).}
MOTSP scores mask shape only, the mean IoU over matched pairs $\mathcal{M}_t$:
\begin{equation}
\mathrm{MOTSP}=\frac{\sum_t \sum_{(i,j)\in\mathcal{M}_t}\mathrm{IoU}(i,j)}{\sum_t |\mathcal{M}_t|}.
\end{equation}
By averaging over already-matched pairs, it reads boundary quality free of any identity error.

\para{Identity Switches (IDS).}
IDS is the raw identity swaps, the ground-truth masks whose matched prediction carries a different ID from the frame before:
\begin{equation}
  \text{IDS} = \bigl\lvert\{m\in M \mid c^{-1}(m)\neq\emptyset \,\land\, pred(m)\neq\emptyset \,\land\, id_{c^{-1}(m)}\neq id_{c^{-1}(pred(m))}\}\bigr\rvert.
\end{equation}
It is the error that MOTSA and sMOTSA below fold into a single accuracy number.

\para{Multi-Object Tracking and Segmentation Accuracy (MOTSA).}
MOTSA combines correct matches, false positives and identity switches:
\begin{equation}
    \text{MOTSA} = \frac{\sum_t (|TP_t| - |FP_t| - |IDS_t|)}{\sum_t |GT_t|}.
\end{equation}

\para{Soft MOTSA (sMOTSA).}
IoU-weighted MOTSA that softens TP by its mask IoU rather than counting it whole:
\begin{equation}
    \text{sMOTSA} = \frac{\sum_t (\sum_{(i,j)\in\mathcal{M}_t}\mathrm{IoU}(i,j) - |\mathrm{FP}_t| - |\mathrm{IDS}_t|)}{\sum_t |\mathrm{GT}_t|},
\end{equation}
so a loose but correct mask no longer scores the same as a tight one.

\para{Video Panoptic Quality (VPQ).}
VPQ carries panoptic quality into time, scoring over $k$-frame windows. For each segment it computes IoU between every ground-truth and predicted trajectory pair, counts a TP at IoU$>0.5$ and treats FP/FN as usual. Then

\begin{equation}
    \text{VPQ}_k = \frac{1}{N_{classes}} \sum_{c} \frac{\sum_{(u,\hat{u}) \in T_P^c} \mathrm{IoU}(u,\hat{u})}{|T_P^c| + \tfrac{1}{2}|F_P^c| + \tfrac{1}{2}|F_N^c|}.
\end{equation}
At $k=0$ this collapses to image-level PQ. For $k>0$ any identity switch inside the window costs an FP/FN, so VPQ gains the explicit temporal sensitivity that frame-level mIoU lacks.

\noindent
\revise{\emph{Decomposition and thing/stuff diagnostic value.} PQ factorizes as a Segmentation Quality (SQ) $\times$ Recognition Quality (RQ) product:
\begin{equation}
\text{PQ} = \underbrace{\frac{\sum_{(u,\hat{u}) \in T_P} \mathrm{IoU}(u,\hat{u})}{|T_P|}}_{\text{SQ}}\;\times\;\underbrace{\frac{|T_P|}{|T_P|+\tfrac{1}{2}|F_P|+\tfrac{1}{2}|F_N|}}_{\text{RQ}},
\end{equation}
where low SQ signals poor mask boundaries and low RQ signals recognition/matching failures. Accordingly, VPS benchmarks report VPQ separately on thing and stuff, with $\text{VPQ}_{\text{thing}}\!-\!\text{VPQ}_{\text{stuff}}$ quantifying the thing--stuff imbalance (\secref{sec:method}), diagnostic neither mIoU (VSS) nor AP (VIS) offer.}

\para{Task-wise Metric Selection Guidelines.} \revise{Each task pairs a primary score with a diagnostic one, so capacity and failure modes both show up. \textbf{VSS}: mIoU $+$ $\text{mVC}_n$ ($n\!=\!8,16$) $+$ FPS. \textbf{VIS}: AP $+$ $\text{AR}_{10}$ $+$ STQ-AQ on occlusion-heavy datasets such as OVIS~\cite{Qi2021OccludedVI}. \textbf{VPS}: VPQ split into thing/stuff $+$ PQ$=$SQ$\times$RQ factorisation $+$ multiple window sizes $k$. \textbf{VTS}: sMOTSA $+$ IDS $+$ MOTSP. \textbf{OVVS}: novel-vs.-base bucketing, because hiding the split makes open-set evaluation ambiguous. }

\section{Performance Comparison}\label{sec:experiments}
{We compare the surveyed VSP methods on standard benchmarks, focusing on VSS, VIS and VPS, where the evaluation protocols are mature.}

\begin{table}[!t]
    \centering
    \begin{minipage}[c]{0.48\linewidth}
        \centering
        \scriptsize
        \setlength{\tabcolsep}{2.6mm}
        \refstepcounter{table}\label{tab:vss_cityscapes}
        {\textbf{Table~\thetable.} {\textbf{Quantitative VSS results on the Cityscapes val set \cite{cordts2016cityscapes} in terms of mIoU and FPS.}}\par}
        \vspace{0.35em}
        \renewcommand{\arraystretch}{0.45}
        \resizebox{\linewidth}{!}{%
        \begin{tabular}{lcccc}
            \toprule
            \textbf{Methods} & \textbf{Backbones} & \textbf{mIoU} & \textbf{FPS} \\
            \midrule
            DVSNet \cite{xu2018dynamic} & ResNet-101 &  70.4 & 19.8 \\
            LVS \cite{li2018low} & ResNet-101 & 76.8 & 5.8 \\
            GRFP(5) \cite{nilsson2018semantic} & ResNet-101 & 69.4 & 3.3 \\
            Accel \cite{Jain2018AccelAC} & ResNet-101 & 75.5 & 1.1 \\
            TDNet \cite{Hu2020TemporallyDN} & ResNet-50 & 79.9 & 5.6 \\
            ETC \cite{liu2020efficient} & ResNet-18 & 73.1 & 9.5 \\
            DepthMix \cite{Hoyer2020ThreeWT} & ResNet-101 & 71.2 & 1.9 \\
            CIRKD \cite{Yang2022CrossImageRK} & ResNet-18 & 74.7 & - \\
            MRCFA \cite{sun2022mining} & MiT-B1 & 75.1 & 21.5 \\
            CFFM \cite{Sun2022CoarsetoFineFM} & MiT-B1 & 75.1 & 23.6 \\
            MPVSS \cite{weng2023mask} & Swin-L & 81.6 & 7.2 \\
            BLO \cite{yang2023pruning} & TopFormer & 74.7 & \textbf{30.8} \\
            SSLTM \cite{Lao2023SimultaneouslySA} & ResNet-50 & 79.7 & - \\
            VPSeg \cite{Guo2024VanishingPointGuidedVS}& MiT-B3 & \textbf{82.5} & - \\
            CFFM++ \cite{Sun2022LearningLA} & MiT-B1 & 78.7 & 20.4 \\
            TV3S \cite{TV3S2025Hesham} & MiT-B1 & 75.6 & - \\
            \bottomrule
        \end{tabular}}
    \end{minipage}\hfill
    \begin{minipage}[c]{0.48\linewidth}
        \centering
        \scriptsize
        \setlength{\tabcolsep}{2.2mm}
        \refstepcounter{table}\label{tab:vss_vspw}
        {\textbf{Table~\thetable.} {\textbf{Benchmark results for VSS on the VSPW validation dataset \cite{miao2021vspw}.} Results are reported with mIoU, mVC and FPS. Methods in \gray{gray} use open-vocabulary supervision.}\par}
        \vspace{0.35em}
        \renewcommand{\arraystretch}{1.1}
        \resizebox{\linewidth}{!}{%
        \begin{tabular}{lccccc}
            \toprule
            \textbf{Methods} & \textbf{Backbones} & \textbf{mIoU} & \textbf{$\text{mVC}_{8}$} & \textbf{$\text{mVC}_{16}$} & \textbf{FPS} \\
            \midrule
            CFFM \cite{Sun2022CoarsetoFineFM} & MiT-B5 &  49.3 & 90.8 & 87.1 & 11.3 \\
            CFFM++ \cite{Sun2022LearningLA} & MiT-B5 & 50.1 & 90.8 & 87.4 & 10.4 \\
            MRCFA \cite{sun2022mining} & MiT-B5 & 49.9 & 90.9 & 87.4 & -- \\
            TCB \cite{miao2021vspw} & ResNet-101 & 37.5 & 87.0 & 82.1 & 10.0 \\
            Video K-Net \cite{li2022video} & ResNet-101 & 38.0 & 87.2 & 82.3 & -- \\
            MPVSS \cite{weng2023mask} & ResNet-101 & 38.8 & 84.8 & 79.6 & -- \\
            TV3S \cite{TV3S2025Hesham} & MiT-B5 & 49.8 & 91.7 & \textbf{88.7} & -- \\
            TubeFormer \cite{kim2022tubeformer} & Axial-ResNet & \textbf{63.2} & \textbf{92.1} & 87.9 & -- \\ \midrule
            \gray{OV2VSS \cite{li2024towards}} & \gray{ResNet-101} & 17.2 & \gray{-} & \gray{-} & \gray{--} \\
            \bottomrule
        \end{tabular}}
    \end{minipage}
    \vspace{-10pt}
\end{table}

\subsection{VSS Benchmark}
\revise{\paraem{Setup.} On Cityscapes~\cite{cordts2016cityscapes}, we report mIoU and FPS. For \tabref{tab:vss_cityscapes} FPS values are taken from the unified benchmarking in~\cite{Wang2021ASO} (single NVIDIA RTX~2080~Ti), with a few entries borrowed from original papers when no implementation is public. For \tabref{tab:vss_vspw} (VSPW) FPS values follow CFFM++~\cite{Sun2022LearningLA} on a single RTX~6000 (24G) at $480\times853$. Entries marked ``--'' are not reported under these protocols.}

\subsubsection{Results}
\revise{\tabref{tab:vss_cityscapes} and \tabref{tab:vss_vspw} suggest that VSS gains come from different mechanisms under different constraints. On Cityscapes, methods that inject strong geometric or scene priors, such as VPSeg~\cite{Guo2024VanishingPointGuidedVS}, tend to improve peak accuracy because the urban layout is relatively structured. In contrast, BLO~\cite{yang2023pruning} and the CFFM family~\cite{Sun2022CoarsetoFineFM,Sun2022LearningLA} perform well on the efficiency side because they avoid recomputing dense features for every frame, either by pruning redundant keyframes or reusing temporally stable features. This explains why the best accuracy and speed are not achieved by the same design, since one exploits stronger spatial priors, while the other exploits temporal redundancy. On VSPW, TubeFormer~\cite{kim2022tubeformer} benefits from clip-level tube modeling, which directly aligns with the dataset's long and diverse videos, whereas TV3S~\cite{TV3S2025Hesham} instead improves temporal consistency through state-space propagation. The remaining gap of OV2VSS~\cite{li2024towards} shows that open-vocabulary VSS is currently limited more by category recognition than by temporal smoothing.}

\begin{figure*}[!t]
    \centering
    \subfloat[Cityscapes val (bubble size $\propto$ FPS).]{\includegraphics[width=0.46\linewidth]{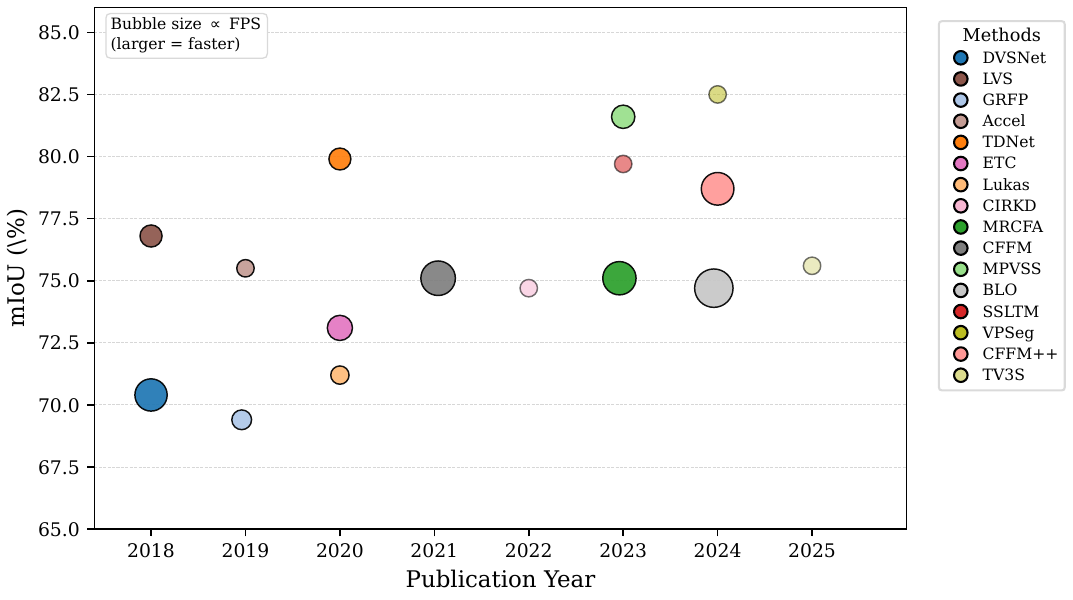}}\hfill
    \subfloat[VSPW val (bubble size $\propto$ $\mathrm{mVC}_{16}$).]{\includegraphics[width=0.46\linewidth]{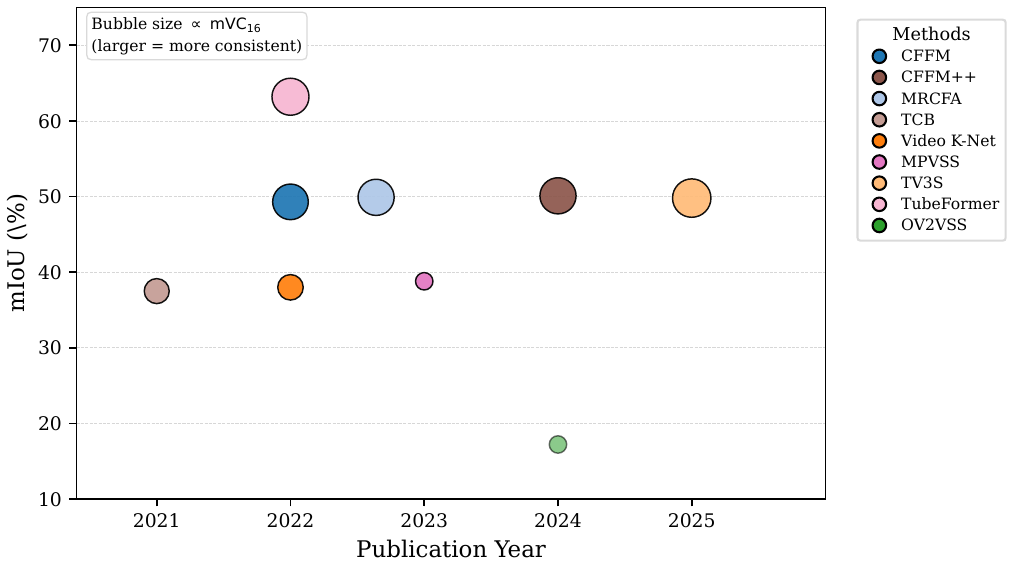}}
    \vspace{-10pt}
    \caption{\revise{\textbf{Performance evolution of VSS methods on Cityscapes and VSPW.} Color encodes method; bubble size $=$ FPS in (a), $\mathrm{mVC}_{16}$ in (b).}}
    \vspace{-15pt}
    \label{fig:vss_evolution}
\end{figure*}

\revise{\paraem{Trend Analysis.}
\figref{fig:vss_evolution} extends the same mechanism-level split across time. Cityscapes rewards ever-stronger per-frame parsing and urban-scene priors, whereas the largest efficiency gains come from designs that cut temporal redundancy rather than enlarge the backbone. VSPW shifts the pressure toward long-range consistency and scene diversity, which favors tube-based and state-space methods, since they preserve information over longer clips. The open-vocabulary point stays separated from the supervised trajectory, a sign that adding temporal context alone is insufficient when the semantic classifier cannot reliably recognize unseen or rare categories.}

\subsection{VIS Benchmark}
\revise{\paraem{Setup.} We follow~\cite{Yang2019VideoIS} to report spatio-temporal-IoU--based AP and AR on the YouTube-VIS validation set. In \tabref{tab:vis_youtube}, FPS values come from each method's original paper or from third-party comparisons under identical settings. We annotate hardware differences via superscripts: $\dagger$~V100, $\ddagger$~RTX~2080~Ti, $\S$~GTX~1080~Ti, $\P$~A100. FPS comparisons should be read with this caveat.}

\subsubsection{Results}
\revise{\tabref{tab:vis_youtube} suggests recent VIS progress is driven less by larger backbones than by better instance-level temporal reasoning. The strongest supervised methods share a query-based formulation, where learnable queries provide a stable slot for each object, so segmentation and association are solved jointly. CTVIS~\cite{ying2023ctvis} improves this direction further by making query representations more discriminative across frames, which directly targets identity switches under occlusion. The AR gap still shows that retrieving multiple plausible instances is easier than ranking the correct long-term association first, so VIS remains sensitive to query competition and identity ambiguity. Open-vocabulary methods lag behind even with comparable backbones, indicating that their main weakness is semantic grounding for unseen categories rather than mask generation alone.}

\begin{table}[!t]
    \centering
    \setlength{\tabcolsep}{3mm} 
    \caption{{\textbf{Evaluation of VIS methods on the YouTube-VIS validation set \cite{Yang2019VideoIS} in terms of accuracy and efficiency.} Methods in \gray{gray} use open-vocabulary supervision. FPS superscripts ($\dagger/\ddagger/\S/\P$) denote the GPU used; see the Setup paragraph of this benchmark for the legend. FPS entries without a superscript come from papers that do not disclose the GPU.}}
    \vspace{-10pt}
    \label{tab:vis_youtube}
    \resizebox{\linewidth}{!}{
    \begin{tabular}{lccccccc}
        \toprule
        \textbf{Methods} & \textbf{Backbones} & \textbf{AP50} & \textbf{AP75} & \textbf{AR1} & \textbf{AR10} & \textbf{AP} & \textbf{FPS} \\
        \midrule
        MaskTrack R-CNN \cite{Yang2019VideoIS} & ResNet-50 & 51.1 & 32.6 & 31.0 & 35.5 & 30.3 & 20.0$^\dagger$ \\
        STEm-Seg \cite{athar2020stem} & ResNet-50 & 50.7 & 33.5 & 31.6 & 37.1 & 30.6 & 7.0$^\dagger$ \\
        fIRN \cite{Liu2021WeaklySI} & ResNet-50 & 27.2 & 6.2 & 12.3 & 13.6 & 10.5 & -- \\
        CrossVIS \cite{yang2021crossover} & ResNet-50 & 56.8 & 38.9 & 35.6 & 40.7 & 39.8 & 39.8$^\dagger$ \\
        SemiTrack \cite{Fu2021LearningTT} & ResNet-50 & 61.1 & 39.8 & 36.9 & 44.5 & 38.3 & -- \\
        MaskProp \cite{bertasius2020classifying} & STSN-ResNeXt-101-64x4d & - & 51.2 & 44.0 & 52.6 & 46.6 & -- \\
        CompFeat \cite{fu2021compfeat} & ResNet-50 & 56.0 & 38.6 & 33.1 & 40.3 & 35.3 & -- \\
        TraDeS \cite{Wu2021TrackTD} & DLA-34 & 52.6 & 32.8 & 30.1 & 29.1 & 31.6 & -- \\
        SG-Net \cite{liu2021sg} & ResNet-50 & 56.1 & 36.8 & 35.8 & 40.8 & 34.8 & -- \\
        VisTR \cite{wang2021end} & ResNet-50 & 59.8 & 36.9 & 37.2 & 42.4 & 36.2 & 69.9$^\dagger$ \\
        Propose-Reduce \cite{lin2021video} & ResNet-50 & 63.0 & 43.8 & 41.1 & 49.7 & 40.4 & -- \\
        TeViT \cite{yang2022temporally} & MsgShifT & 71.3 & 51.6 & 44.9 & 54.3 & 46.6 & 68.9$^\dagger$ \\
        VISOLO \cite{han2022visolo} & ResNet-50 & 56.3 & 43.7 & 35.7 & 42.5 & 38.6 & 40.0$^\ddagger$ \\
        SeqFormer \cite{wu2022seqformer} & ResNet-50 & 69.8 & 51.8 & 45.5 & 54.8 & 47.4 & 72.3$^\dagger$ \\
        IDOL \cite{wu2022defense} & ResNet-50 & 74.0 & 52.9 & 47.7 & 58.7 & 49.5 & 30.6 \\
        STC \cite{Jiang2022STCSC} & ResNet-50 & 57.2 & 38.6 & 36.9 & 44.5 & 36.7 & 40.3$^\ddagger$ \\
        MinVIS \cite{huang2022minvis} & ResNet-50 & 69.0 & 52.1 & 45.7 & 55.7 & 47.4 & -- \\
        GenVIS \cite{heo2023generalized} & ResNet-50 & 72.0 & 57.8 & 49.5 & 60.0 & 51.3 & 35.6$^\P$ \\
        VideoCutLER \cite{Wang2023VideoCutLERSS} & ResNet-50 & 50.7 & 24.2 & - & 42.4 & 26.0 & -- \\
        DVIS \cite{zhang2023dvis} & ResNet-50 & 76.5 & 58.2 & 47.4 & 60.4 & 52.6 & 18.7$^\S$ \\
        CTVIS \cite{ying2023ctvis} & ResNet-50 & \textbf{78.2} & \textbf{59.1} & \textbf{51.9} & \textbf{63.2} & \textbf{55.1} & -- \\ \midrule
        \gray{OV2Seg+ \cite{wang2024ov}} & \gray{ResNet-50} & \gray{-} & \gray{-} & \gray{-} & \gray{-} & \gray{32.9} & \gray{--} \\
        \gray{OVFormer \cite{fang2024unified}} & \gray{ResNet-50} & \gray{-} & \gray{-} & \gray{-} & \gray{-} & \gray{34.8} & \gray{--} \\
        \gray{CLIP-VIS \cite{zhu2025clip}} & \gray{ResNet-50} & \gray{-} & \gray{-} & \gray{-} & \gray{-} & \gray{32.2} & \gray{--} \\
        \bottomrule
    \end{tabular}}
    \vspace{-20pt}
\end{table}

\begin{figure}[!t]
    \centering
    \subfloat[YouTube-VIS val: bubble size $\propto$ $\mathrm{AR}_{10}$.\label{fig:vis_evolution}]{\includegraphics[width=0.46\linewidth]{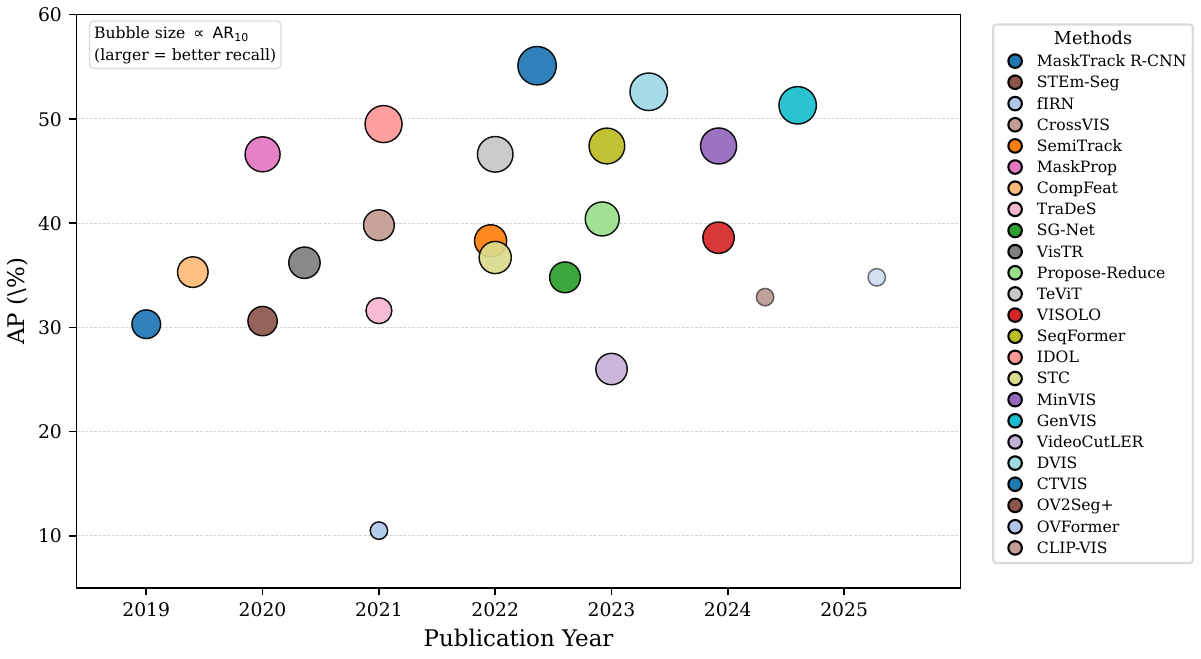}}\hfill
    \subfloat[KITTI-MOTS Cars val: bubble size $\propto$ $1/\text{IDS}$.\label{fig:vts_evolution}]{\includegraphics[width=0.46\linewidth]{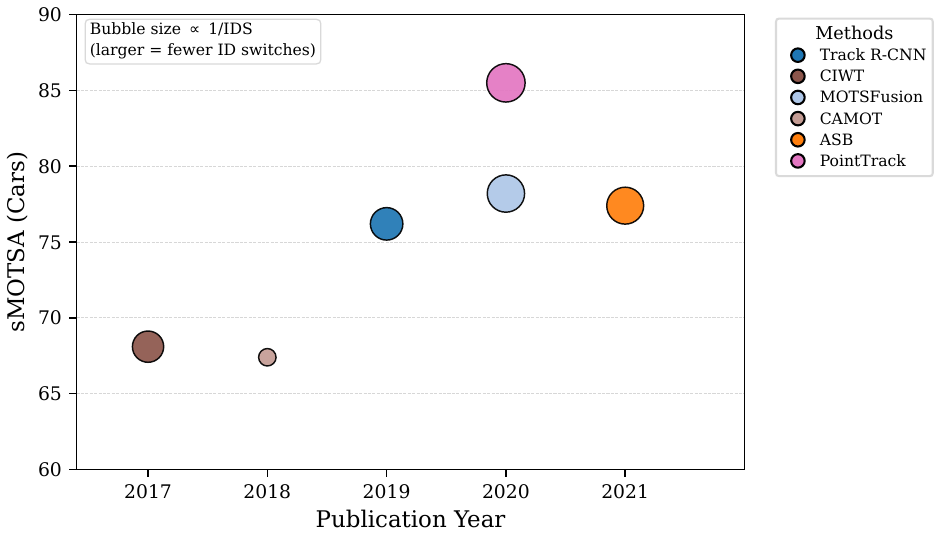}}
    \vspace{-10pt}
    \caption{\revise{\textbf{Performance evolution of VIS and VTS methods.} Left: VIS on YouTube-VIS, bubble $=\mathrm{AR}_{10}$. Right: VTS on KITTI-MOTS Cars, bubble $=1/\text{IDS}$ (larger $=$ more stable).}}
    \vspace{-7pt}
    \label{fig:vis_vts_evolution}
\end{figure}

\revise{\paraem{Trend Analysis.}
\figref{fig:vis_evolution} clarifies architectural transition. Early tracking-by-detection systems inherit detection errors before association as they depend on per-frame proposals. The later query-based wave improves because it treats instances as persistent set elements across the clip, allowing masks and identities to be refined together. This explains improvement in both precision and recall: better queries reduce duplicate or fragmented tracks while recovering objects. The lower open-vocabulary cluster marks a different bottleneck: once language-conditioned category assignment is introduced, recognition uncertainty becomes a limiting factor even when temporal association is strong.}

\subsection{VPS Benchmark}
\revise{\paraem{Setup.} We report VPQ and STQ on Cityscapes-VPS, KITTI-STEP and VIPSeg. FPS values in \tabref{tab:vps} aggregate original papers and cross-method comparisons: VPSNet-FuseTrack from~\cite{ye2022hybridtracker} (Cityscapes-VPS val), ViP-DeepLab approximated as $>$400~ms/frame ($\approx$2.5~FPS) by~\cite{petrovai2022time}, and Video K-Net from~\cite{cheng2023tracking}. Hardware differs across these works (\eg, V100 vs.\ Titan X), so absolute FPS is indicative rather than strictly comparable.}

\subsubsection{Results}
\revise{\tabref{tab:vps} shows that VPS performance is highly coupled to dataset structure. PolyphonicFormer~\cite{yuan2022polyphonicformer} works well on Cityscapes-VPS because its unified query design and depth-aware reasoning fit structured driving scenes, while Video K-Net~\cite{li2022video} and TubeLink~\cite{li2023tube} are stronger on benchmarks that reward tube-level or kernel-based temporal grouping. No single method leads everywhere, which shows that VPS designs are still tuned to a particular balance between thing tracking, stuff parsing and temporal aggregation. When the benchmark moves from compact urban labels to broader in-the-wild vocabularies, the key difficulty shifts from panoptic fusion to long-tail recognition and stable thing--stuff allocation. Latency remains a secondary reporting axis, which suggests that the field has prioritized unified representation quality before real-time deployment.}

\begin{table*}[!t]
    \vspace{-5pt}
    \centering
    \setlength{\tabcolsep}{1mm} 
    \caption{{\textbf{Benchmark results on VPS validation datasets} (VPQ, STQ, FPS).}}
    \vspace{-10pt}
    \label{tab:vps}
    \resizebox{\textwidth}{!}{%
    \begin{tabular}{lccccc}
        \toprule
        \textbf{Methods} & \textbf{Backbones} & \textbf{Cityscapes-VPS \cite{Kim2020VideoPS} (VPQ)} & \textbf{KITTI-STEP \cite{Weber2021STEPSA} (STQ)} & \textbf{VIPSeg \cite{Miao2022LargescaleVP} (STQ)} & \textbf{FPS} \\
        \midrule
        VPSNet \cite{Kim2020VideoPS} & ResNet-50 & 57.4 & - & - & 1.3 \\
        VIP-Deeplab \cite{Qiao2020ViPDeepLabLV} & ResNet-50 & 63.1 & - & - & $\sim$2.5 \\
        SiamTrack \cite{woo2021learning} & ResNet-50 & 57.8 & - & - & -- \\
        Clip-PanoFCN \cite{Miao2022LargescaleVP} & ResNet-50 & - & - & 31.5 & -- \\
        PolyphonicFormer \cite{yuan2022polyphonicformer} & ResNet-50 & \textbf{65.4} & - & - & -- \\
        TubeFormer \cite{kim2022tubeformer} & Axial-ResNet & - & 70.0 & - & -- \\
        SLOT-VPS \cite{zhou2022slot} & Swin-L & 63.7 & - & - & -- \\
        TubeLink \cite{li2023tube} & Swin-B & - & 72.2 & \textbf{49.4} & -- \\
        Video K-Net \cite{li2022video} & Swin-B & 62.2 & \textbf{73.0} & 46.3 & 4.9 \\
        \bottomrule
    \end{tabular}}
    \vspace{-15pt}
\end{table*}

\revise{\paraem{Trend Analysis.}
The VPS trend splits into two regimes. Compact driving benchmarks improve quickly because their scene layout, class set, and motion patterns are regular enough for stronger query or depth-aware fusion to absorb most errors. VIPSeg remains harder because generic videos expose long-tail categories, diverse scenes, and less predictable thing--stuff boundaries. This implies that future VPS gains will likely come less from another panoptic fusion head alone and more from better category priors, data-efficient long-tail learning, and temporal mechanisms that can keep thing identities stable without degrading stuff consistency.}

\subsection{VTS Benchmark}
\revise{\paraem{Setup.} We report sMOTSA, MOTSA and IDS on KITTI-MOTS val in \tabref{tab:vts}. The FPS values for Track R-CNN, MOTSFusion and PointTrack are converted from the per-frame timings reported under the unified single-GPU protocol of PointTrack~\cite{xu2020segment} (Tab.~2, end-to-end detection+segmentation+tracking on KITTI-MOTS val); the GPU model is not disclosed in the source paper, so absolute FPS is indicative.}

\subsubsection{Results}
\revise{\tabref{tab:vts} identifies PointTrack~\cite{xu2020segment} as the clearest example that identity representation can matter more than detector choice in VTS. Since the compared methods largely share the same detection source, the performance gap is better explained by how each method represents and matches instances. PointTrack's point-based embedding preserves object geometry and deformation cues more directly than ROI-level appearance features, which makes associations more stable under motion and occlusion. The remaining gap between cars and pedestrians indicates a different limit, since small, articulated and frequently occluded objects still challenge strong association designs. Thus, VTS progress is driven by identity-stable representations rather than by mask quality alone.}

\begin{table}[!t]
    \centering
    \setlength{\tabcolsep}{2.5mm}
    \caption{{\textbf{Evaluation of VTS methods on KITTI-MOTS val~\cite{voigtlaender2019mots}.}}}
    \vspace{-10pt}
    \label{tab:vts}
    \resizebox{\linewidth}{!}{%
    \begin{tabular}{lc|ccc|ccc|c}
        \toprule
        \multirow{2}{*}{\textbf{Methods}} 
        & \multirow{2}{*}{\textbf{Detectors}} 
        & \multicolumn{3}{c|}{\textbf{Cars}} 
        & \multicolumn{3}{c|}{\textbf{Pedestrians}}
        & \multirow{2}{*}{\textbf{FPS}}
        \\ \cmidrule(lr){3-5} \cmidrule(lr){6-8}
        &  
        & \textbf{sMOTSA} & \textbf{MOTSA} & \textbf{IDS} 
        & \textbf{sMOTSA} & \textbf{MOTSA} & \textbf{IDS} 
        & \\ \midrule
        Track R-CNN \cite{voigtlaender2019mots} & Track R-CNN & 76.2 & 87.8 & 93 & 46.8 & 65.1 & 78 & 2.0 \\
        CIWT \cite{osep2017combined} & Track R-CNN & 68.1 & 79.4 & 106 & 42.9 & 61.0 & 42 & -- \\
        MOTSFusion \cite{luiten2020track} & Track R-CNN & 78.2 & 90.0 & 36 & 50.1 & 68.0 & 34 & 1.2 \\
        CAMOT \cite{ovsep2018track} & Track R-CNN & 67.4 & 78.6 & 220 & 39.5 & 57.6 & 131 & -- \\
        ASB \cite{choudhuri2021assignment} & Track R-CNN & 77.4 & 89.6 & 41 & 48.9 & 66.7 & 28 & -- \\
        PointTrack \cite{xu2020segment} & Track R-CNN & \textbf{85.5} & \textbf{94.9} & \textbf{22} & \textbf{62.4} & \textbf{77.3} & \textbf{19} & \textbf{22.2} \\
        \bottomrule
    \end{tabular}}
    \vspace{-15pt}
\end{table}

\revise{\paraem{Trend Analysis.}
\figref{fig:vts_evolution} shows that segmentation quality and tracking stability improve together only when the association representation becomes discriminative. Earlier matching heuristics can produce acceptable masks while still causing identity drift, because they optimize frame-level overlap more than long-horizon identity continuity. PointTrack changes the trend by embedding instances in a geometry-aware space, reducing the dependence on brittle appearance matching. This suggests that future VTS gains should focus on representations that remain stable through occlusion, deformation, and re-entry, rather than only improving per-frame segmenter.}

\subsection{OVVS Benchmark}
\revise{\tabref{tab:vss_vspw} and \tabref{tab:vis_youtube} reveal a consistent semantic bottleneck. Supervised methods possess strong mask decoders and temporal association modules, while open-vocabulary variants additionally align pixels, object identities, and language categories that are rare or unseen during training. This extra grounding step explains why the gap persists across both VSS and VIS even when backbones are comparable. The problem is therefore not simply weaker segmentation or tracking but the difficulty of transferring image-language semantics into temporally coherent video masks. Closing this gap likely requires better video-language supervision, rare-category calibration, and mechanisms that keep category identity stable across frames, reinforcing the open-world direction discussed in \secref{sec:cross-task} and \secref{sec:future}.}

\section{Future Directions}\label{sec:future}
\para{Open-World Video Scene Parsing.} \revise{\textit{Bottleneck.} \revise{Closed-vocabulary VSP saturates in-distribution but collapses on rare or unseen categories: \figref{fig:vis_evolution} shows OV2Seg+/OVFormer/CLIP-VIS trailing supervised SOTA by 20--23 AP, and \tabref{tab:vss_vspw} shows OV2VSS trailing TubeFormer by about 46 mIoU on VSPW.}
\textit{Technical challenge.} Pixel-level mask quality (localization) must be decoupled from open-set recognition (long-tail, prior-driven). CLIP-based heads inherit image-level biases and degrade on small or temporally inconsistent objects, while continual learning risks forgetting base classes as the taxonomy grows.
\textit{Candidate directions.} (i) decoupled architectures routing mask proposals to frozen VLMs for classification; (ii) memory-efficient continual segmentation with mask-token rehearsal; (iii) bootstrapping novel-class supervision from web-scale image--text pairs aligned temporally via cross-frame mask consistency.
Representative recent works: \cite{wang2021unidentified,wu2024towards,guo2024videosam,thawakar2025video,xue2024learning,tang2025towards,kalluri2024open}.}

\para{Unified Video Scene Parsing.} \revise{\textit{Bottleneck.} Maintaining four model families (VSS/VIS/VPS/VTS) on disjoint datasets with task-specific heads breeds redundant engineering, inconsistent temporal-modeling choices, and no clean way to transfer one task's progress to another without re-implementation.
\textit{Technical challenge.} Unification has to reconcile heterogeneous supervision (semantic/instance/ panoptic/trajectory) and metrics (mIoU/AP/VPQ-STQ/sMOTSA-IDS) without one objective dominating, while staying competitive on every leaderboard under a shared parameter budget.
\textit{Candidate directions.} (i) shared mask-tokens-as-queries decoupling ``what to segment'' from ``how to label''; (ii) task-conditioning via natural-language prompts; (iii) cross-task curricula that exploit data-rich VSS to bootstrap data-scarce VPS/VTS.
Representative recent works: \cite{athar2023burst,li2024omg,chen2024general,zhang2024omg,song2024ba,yuan2025instruction,zhang2025dvis++,yan2024referred}.}

\para{Multimodal Fusion for Segmentation.} \revise{\textit{Bottleneck.} Pure-RGB VSP fails when a target is camouflaged or in low light, when boundaries are ambiguous by appearance (glass, water), or when identity must persist across long appearance-changing occlusions.
\textit{Technical challenge.} Effective fusion must address modality misalignment in space (depth--RGB calibration drift), in time (audio--visual synchronization jitter) and in semantics (sparse text vs.\ dense pixels), without letting the dominant RGB stream suppress weaker but disambiguating signals.
\textit{Candidate directions.} (i) late-stage cross-modal attention with learned modality dropout; (ii) self-supervised cross-modal pre-training where one modality predicts another; (iii) modality-adaptive routing where a per-pixel gate selects the locally most reliable modality.
Representative recent works: \cite{voigtlaender2023connecting,wang2023visionllm,li2024monkey,lu2024unified,zou2023generalized,zhang2023meta,baldassini2024makes,srinivasan2022climb}.}

\para{Visual Reasoning for Segmentation.} \revise{\textit{Bottleneck.} As our failure-mode analysis (\secref{sec:failure}) shows, SOTA VSP still fails on prolonged occlusion, identity switches in crowded scenes and rare fine-grained categories, failures rooted not in pixel capacity but in the absence of explicit relational reasoning over time and space.
\textit{Technical challenge.} Endowing segmentation with reasoning needs one representation that supports both fast dense per-pixel prediction (sub-100\,ms) and slow structured relational inference (object permanence, causal interaction), two historically incompatible regimes.
\textit{Candidate directions.} (i) two-stream designs pairing a fast mask decoder with a slow scene-graph reasoner; (ii) physics-informed dynamics that hallucinate occluded masks via object-permanence priors; (iii) language-mediated reasoning where a VLM emits a textual scene description grounded by a downstream mask network.
Representative recent works: \cite{zheng2024villa,yan2024visa,bai2024one,gong2025devil}.}

\para{Generative Segmentation.} \revise{\textit{Bottleneck.} Discriminative segmenters learn one-to-one pixels to mask mapping and hence cannot model uncertainty (multiple plausible masks under occlusion) or hallucinate masks for transiently invisible objects, both of which are central to video.
\textit{Technical challenge.} Adapting diffusion/generative priors requires (i) reconciling iterative stochastic sampling with deterministic real-time demands, (ii) preserving cross-frame coherence under per-frame stochasticity, and (iii) producing calibrated uncertainty rather than visually plausible but inconsistent samples.
\textit{Candidate directions.} (i) deterministic distillation of diffusion mask priors into a single forward pass; (ii) joint frame-level diffusion with cross-frame consistency losses; (iii) using generative models as auxiliary supervision (mask-completion pretext) rather than as the primary inference engine.
Representative recent works: \cite{chen2023generalist,gu2024diffusioninst,wang2023towards,qi2024unigs,li2023momentdiff}.}

\para{Efficient Video Understanding.} \revise{\textit{Bottleneck.} As \figref{fig:vss_evolution} shows, accuracy and FPS gains have evolved \emph{orthogonally}: top-accuracy methods (TubeFormer, MRCFA) operate at single-digit FPS, while real-time methods (DVSNet, MPVSS) sacrifice accuracy for more throughput, so no method dominates both axes.
\textit{Technical challenge.} Three coupled factors bound the frontier: (i) per-frame attention is $O(N^{2})$ in spatial tokens; (ii) clip-level temporal attention multiplies this by the temporal window; (iii) on edge devices memory bandwidth, not FLOPs, governs wall-clock time, so naive FLOP cuts need not translate into speedups.
\textit{Candidate directions.} (i) sparse/linear attention exploiting cross-frame redundancy; (ii) motion-saliency-conditioned dynamic token pruning; (iii) keyframe$+$propagation hybrids that re-use heavy features over $K$ frames; (iv) hardware-aware NAS co-optimizing target-accelerator latency.
Representative recent works: \cite{lin2019tsm,mehta2021mobilevit,wu2022memvit,zhang2023faster,xu2023pidnet,gao2023rethinking}.}

\para{Large Language Model-based Segmentation.} \revise{\textit{Bottleneck.} LLM-grounded segmenters are mostly evaluated on still images, so on video three problems surface: (i) frame-by-frame inconsistency of textual reasoning; (ii) prohibitive inference cost (hundreds of ms/frame); (iii) poor long-horizon coherence, since most VLMs are trained on $\le$8-frame clips.
\textit{Technical challenge.} The open question is how to ground a slowly-changing language plan (\eg, ``segment the truck overtaking the cyclist'') onto fast-changing pixel masks, preserving identity through occlusion and reappearance, in a single forward pass.
\textit{Candidate directions.} (i) cache LLM ``intent tokens'' once per shot and reuse them across frames; (ii) chain-of-thought reasoning over compressed temporal memories rather than raw pixels; (iii) frozen-LLM--fast-mask co-design that decouples reasoning frequency from segmentation frequency.
Representative recent works: \cite{wang2024llm,li2025visual,shi2025llmformer,lai2024lisa,xia2024gsva,chen2024internvl,ma2024groma,chen2024sharegpt4video}.}

\section{Conclusion}
{This survey reviews the evolution of VSP from hand-crafted features to deep convolutional, attention-based, and Transformer architectures, which brought consistent gains in segmentation accuracy and temporal coherence. Several problems remain open (principled spatio-temporal fusion, long-horizon consistency, and recognition beyond closed vocabularies), and a clear gap to industrial deployment persists despite sharper datasets and metrics. Future progress will depend on efficiency, robustness to real-world dynamics, and adaptability to unseen concepts. We hope that this consolidation, together with the directions outlined above, provides a useful reference for future work on video scene parsing.}

\begin{acks}
This work was supported by the National Natural Science Foundation of China (No. 62576176). The computational resources were supported by the Supercomputing Center of Nankai University (NKSC).
\end{acks}

{\linespread{0.96}\selectfont
\bibliographystyle{ACM-Reference-Format}
\bibliography{reference}}





\end{document}